\documentclass[]{youtu} 

\usepackage{mathpazo}
\usepackage{graphicx}
\usepackage{natbib}
\usepackage{amssymb} 
\usepackage{multirow}
\title{CoDiEmb: A Collaborative yet Distinct Framework for Unified Representation Learning in Information Retrieval and Semantic Textual Similarity}

\author{Bowen Zhang\textsuperscript{$1$}$^*$, Zixin Song\textsuperscript{$1$}$^*$, Chunquan Chen\textsuperscript{$2$}, Qian-Wen Zhang\textsuperscript{$2$}$\dagger$, Di Yin\textsuperscript{$2$}, Xing Sun\textsuperscript{$2$}}
\vspace{-5mm}
\affiliation{$^1$Tsinghua University\\\textsuperscript{$2$}Tencent Youtu Lab }

\sourcecode{https://github.com/TencentCloudADP/youtu-embedding.git}

\begin{document} 

\abstract{
Learning unified text embeddings that excel across diverse downstream tasks is a central goal in representation learning, yet negative transfer remains a persistent obstacle. This challenge is particularly pronounced when jointly training a single encoder for Information Retrieval (IR) and Semantic Textual Similarity (STS), two essential but fundamentally disparate tasks for which naive co-training typically yields steep performance trade-offs. We argue that resolving this conflict requires systematically decoupling task-specific learning signals throughout the training pipeline. To this end, we introduce CoDiEmb, a unified framework that reconciles the divergent requirements of IR and STS in a collaborative yet distinct manner. CoDiEmb integrates three key innovations for effective joint optimization: (1) Task-specialized objectives paired with a dynamic sampler that forms single-task batches and balances per-task updates, thereby preventing gradient interference. For IR, we employ a contrastive loss with multiple positives and hard negatives, augmented by cross-device sampling. For STS, we adopt order-aware objectives that directly optimize correlation and ranking consistency. (2) A delta-guided model fusion strategy that computes fine-grained merging weights for checkpoints by analyzing each parameter's deviation from its pre-trained initialization, proving more effective than traditional Model Soups. (3) An efficient, single-stage training pipeline that is simple to implement and converges stably. Extensive experiments on 15 standard IR and STS benchmarks across three base encoders validate CoDiEmb. Our results and analysis demonstrate that the framework not only mitigates cross-task trade-offs but also measurably improves the geometric properties of the embedding space.

}

\maketitle
\footnotetext[1]{$^*$Work done during the internship at Tencent Youtu Lab.}
\footnotetext[2]{$\dagger$Corresponding author(s): cowenzhang@tencent.com}

\vspace{-.1em}


\section{Introduction}

Modern Natural Language Processing (NLP) is largely driven by two paradigms: generation and encoding \citep{GRIT-ICLR-2024}. The output of encoder models, known as text embeddings, represents a cornerstone of computational linguistics. Among the myriad applications and benchmarks for text embeddings, Semantic Textual Similarity (STS) and Information Retrieval (IR) stand out as two of the most critical tasks \citep{SimCSE-EMNLP-2021}. STS aims to determine the semantic proximity between two text segments, forming the foundation for techniques such as recommendation systems, text clustering, and content normalization \citep{SimTier-CIKM-2024}. IR, on the other hand, focuses on measuring the relevance between a query and a large document collection, playing a pivotal role in search engines, dialogue platforms, and AI agents \citep{DynamicRAG-2025}.

Motivated by the goal of creating a universal text encoder proficient in both task families, state-of-the-art embedding models commonly co-train on large mixtures of STS and IR datasets using contrastive learning \citep{CMTEB-BGE-SIGIR-2024, NV-Embed-2024}. While straightforward, this practice overlooks the inherent discrepancies between the two task types. Concretely, STS and IR exhibit significant differences in several key aspects:
\begin{itemize}
\item \textbf{Data Structure and Symmetry:} STS tasks typically organize data in triplets $(x_1, x_2, y)$, where the paired texts $x_1$ and $x_2$ are highly symmetric; swapping their positions does not alter the label $y$. In contrast, IR datasets are inherently asymmetric, comprising a set of queries $\{q\}_i$, a large document corpus $\{d\}_j$, and a relevance mapping $\{(q_i, d_j)\}_1^l$ that defines their relationships. During inference, a query $q_i$ is matched against each document in $\{d\}_1^n$, but only the pairs $(q_i, d_j)$ specified in the mapping are considered relevant.

\item \textbf{Semantic Granularity and Text Length:} STS tasks demand fine-grained semantic distinctions, and their training and evaluation sets often feature granular annotation scores. As the definition of semantic similarity becomes more ambiguous with increasing text length \citep{C-STS-EMNLP-2023}, STS sequences are generally short. Conversely, the lengths of queries and documents in IR are highly flexible, with documents frequently spanning hundreds of tokens. As a result, although both tasks leverage cosine similarity for efficient matching, the underlying meaning of the calculation differs: STS prioritizes semantic equivalence, whereas IR leans towards topical or knowledge-level relevance.

\item \textbf{Evaluation Metrics:} The primary metric for STS is Spearman's rank correlation coefficient \citep{Spearman-2005}, which measures the monotonic relationship between predicted and true rankings. The Normalized Discounted Cumulative Gain (nDCG) metric \citep{nDCG-PMLR-2013} used in IR is also list-wise but places greater emphasis on the correctness of top-ranked items. Furthermore, because relevant documents for a given query are typically sparse in IR, nDCG@k is commonly adopted.
\end{itemize}

These discrepancies lead to suboptimal performance when the two tasks are optimized indiscriminately. As we will demonstrate in Section~\ref{sec:experiments}, naively applying an objective function suited for one task, such as InfoNCE Loss \citep{InfoNCE-2018} for IR or CoSENT Loss \citep{CoSENT-2024} for STS, is detrimental to the other. In contrast, our proposed framework, CoDiEmb, strikes a robust balance between IR and STS during co-training, approaching or even surpassing the performance of single-task fine-tuning.

Notably, some cutting-edge research has also observed these performance trade-offs. \citet{Task-Instruction-2022} propose designing distinct instructions for different tasks and prepending them to the input text. While this strategy yields significant improvements, the prior information provided by such instructions is limited and relies entirely on the model's implicit contextual understanding, lacking explicit gradient signals. Jina-embeddings-v3 \citep{Jina-v3-2024} introduces Task LoRA for parameter-level customization, but this necessitates storing a series of adapters. Moreover, if a document appears in $k$ task sets, it would require $k$ distinct embeddings, incurring prohibitive storage costs. NV Embed \citep{NV-Embed-2024} converts all data types into an IR format and constructs a two-stage training pipeline: first fine-tuning on IR datasets with hard negatives, followed by contrastive learning on a mixture of all corpora without hard negatives. This process inevitably discards a large volume of low-score STS data that cannot be formulated into positive pairs. Additionally, as noted in prior work, a coarse-grained contrastive objective is ill-suited for tasks with fine-grained labels \citep{Pcc-tuning-EMNLP-2024, STS-Reg-EMNLP-2024}.

This landscape reveals a pressing need for a unified, effective, and end-to-end solution for the joint optimization of IR and STS. To this end, we present CoDiEmb, a framework that \textbf{\underline{Co}}llaboratively yet \textbf{\underline{Di}}stinctly handles Information Retrieval and Semantic Textual Similarity from multiple perspectives, including loss functions, data sampling, and model fusion.


Specifically, for IR tasks, we design a contrastive loss that supports multiple positives and hard negatives per anchor. This is augmented with cross-device negative sampling, which dramatically expands the pool of comparison candidates, yielding sharper separability. During this process, CoDiEmb's dynamic sampler ensures that, in each iteration, all GPUs draw samples strictly from disjoint subsets of the same data file, thereby providing pure task gradients. For STS tasks, rather than relying on the binary classification-style InfoNCE Loss or approximating the ranking objective by penalizing inverted pairs, we opt for direct optimization of order consistency. Building on the Pearson Loss proposed in Pcc-tuning \citep{Pcc-tuning-EMNLP-2024}, we introduce our modified and adapted KL divergence Loss and PRO Loss \citep{BEQUE-WWW-2024}, which substantially enhance the model's fine-grained semantic discrimination. 

Finally, by analyzing the deviation of fine-tuned parameters from their pre-trained values, we develop an innovative model fusion strategy. Applying this method to checkpoints from different training trajectories yields performance gains beyond those achieved by standard Model Soups \citep{Model-Soups-PMLR-2022}.

In summary, the main contributions of this paper are as follows:
\begin{itemize}

\item We propose CoDiEmb, a framework that enables a model to converge effectively on both IR and STS tasks within a single training stage. CoDiEmb requires no adapter components, and its unified data format is fully compatible with corpora of arbitrary granularity, eliminating the need to discard any samples.

\item We formulate specialized loss functions tailored to the distinct characteristics of IR and STS. In conjunction with our custom dynamic sampler, this approach not only balances per-task iteration counts but also avoids the gradient interference induced by mixed-task batches.

\item By analyzing parameter shifts under different training settings, we devise an effective weighting scheme for ensembling checkpoints. Our method moves beyond conventional model-level fusion to a finer granularity, operating directly on learnable parameters.

\item We conduct extensive experiments with MiniCPM \citep{MiniCPM-COLM-2024}, E5 \citep{M-E5-2024}, and BGE \citep{CMTEB-BGE-SIGIR-2024} on 8 IR and 7 STS benchmarks, thoroughly validating the superiority of CoDiEmb. To further elucidate the underlying principles of our method, we provide a series of theoretical analyses, finding that CoDiEmb's joint optimization strategy effectively mitigates anisotropy \citep{Anisotropy-EMNLP-2019} and over-smoothing \citep{Over-smoothing-2022} in the learned representation space.

\end{itemize}

\section{Methodology}

This section presents CoDiEmb, our end-to-end framework for unified representation learning across STS and IR. We begin in subsection~\ref{sec:data_format} by introducing our task–agnostic data format, explaining its compatibility with inputs of heterogeneous granularity and its extensibility to other tasks. Subsequently, in subsection~\ref{sec:loss_functions}, we provide a detailed exposition of CoDiEmb's specialized loss functions, linking their design to the corresponding evaluation metrics. Building upon this, subsection~\ref{sec:sampler_gpu} describes CoDiEmb's data sampler and multi-device training configuration. Finally, in subsection~\ref{sec:model_fusion}, we introduce our proposed parameter-level model fusion strategy.

\subsection{Unified Data Format}
\label{sec:data_format}

\begin{figure*}[htbp]
\centering
\includegraphics[width=1.0\linewidth]{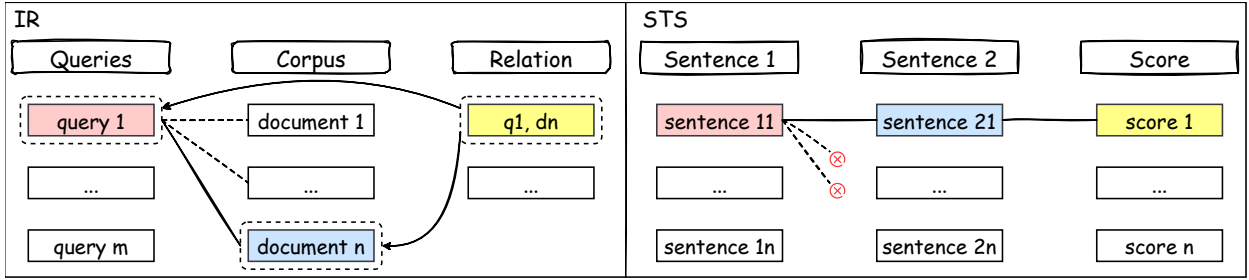}
\caption{A comparison of the dataset structures and evaluation pipelines for typical IR and STS tasks.}
\label{fig:ir_sts_data}
\end{figure*}

IR and STS follow distinct data organization schemes driven by their respective evaluation protocols. As illustrated in Figure~\ref{fig:ir_sts_data}, IR tasks match each query $q_i$ from a set $\{q\}_1^m$ against the entire document corpus $\{d\}_1^n$ to retrieve the top-$k$ most relevant results. Ground-truth relevance is defined by a mapping table $\{(q_i,d_j)\}_1^l$, which typically stores only the identifiers of positive samples. Any pair $(q_i,d_j)$ absent from this mapping is implicitly considered as a negative sample. Among these negative documents, some are more challenging to distinguish from positives and are termed hard negatives. The community has long recognized the critical importance of hard negatives for IR \citep{zhan2021optimizing, SimANS-EMNLP-2022}. Consequently, data mining techniques are often employed to identify a set of hard negatives $\{d^-\}$ for a given query $q$, leading to a data structure of $(q, d^+, \{d^-\})$.

In contrast, pairs within STS tasks, $(x_1, x_2)$, are treated as independent instances. A model directly predicts a score $\hat{y}$ via cosine similarity, and the resulting list of predictions $\{\hat{y}\}_1^n$ is then compared with the ground-truth scores $\{y\}_1^n$ to evaluate rank consistency. Thus, STS data are commonly structured as triplets of $(x_1, x_2, y)$.

To accommodate both data types, CoDiEmb employs a unified format: $(t, q, \{d^+\}_1^m, \{d^-\}_1^n, \{y^+\}_1^m, \{y^-\}_1^n)$. Here, $t$ is a task identifier, which can be specified at the dataset level. Fields absent in the original data are populated with placeholders that are ignored during the forward pass, incurring no additional memory overhead. This integrated data structure is highly extensible. When processing an STS task, we map the triplet $(x_1, x_2, y)$ to the query $q$, the first positive document $d^+_1$, and the first positive score $y^+_1$, respectively. For IR tasks, we populate the query $q$, the positive set $\{d^+\}_1^m$, and the negative set $\{d^-\}_1^n$. If relevance scores are available, they can be stored in the corresponding $\{y\}$ fields.

This extensible format also naturally supports other tasks like classification and clustering. For these tasks, data can be partitioned by labels, allowing for intra-cluster (positive) and inter-cluster (negative) sampling to construct inputs for contrastive learning. The format is also compatible with classifier-head architectures \citep{Sentence-BERT-EMNLP-2019, STS-Reg-EMNLP-2024} by assigning the input text to $q$ and its label to $y^+$. 

Leveraging this unified data structure, CoDiEmb not only standardizes the loading of diverse corpora but also enables the configuration of differentiated loss functions tailored to task characteristics, thereby facilitating multi-granularity training. Although this paper focuses on the joint optimization of IR and STS, the potential of CoDiEmb extends beyond this scope, which we plan to explore in future work.

\subsection{Differentiated Loss Functions}
\label{sec:loss_functions}

As the optimization objective for model training, loss functions have profound impacts on a neural network's performance. A well-designed loss function should closely align with the task's evaluation metrics to provide effective learning signals. The primary metrics for IR and STS are nDCG@k and Spearman's rank correlation coefficient, respectively. Both are non-differentiable ranking metrics and thus cannot be directly used in backpropagation.

For a given query $q$, let the top-$k$ documents retrieved by the model be $\{d_{\theta(1)}, d_{\theta(2)}, ..., d_{\theta(k)}\}$. The nDCG@k is calculated as:
\begin{equation}
\label{eq:ndcg}
    \begin{gathered}
        \text{nDCG@k} = \frac{\text{DCG@k}}{\text{IDCG@k}} \quad 
        \text{DCG@k} = \sum_{i=1}^{k} \frac{\text{rel}_i}{\log_2(i+1)} \\
        \text{IDCG@k} = \sum_{i=1}^{k} \frac{\text{rel}^{\text{ideal}}_i}{\log_2(i+1)}
    \end{gathered}
\end{equation}

Here, $\text{rel}_i$ is the annotated relevance score of the retrieved document $d_{\theta(i)}$ at rank $i$, while $\text{rel}^{\text{ideal}}_i$ is the score of the ideal document at that rank. This formulation reveals that the core objective of nDCG is to place highly relevant documents at the top of the full candidate list. 

\begin{figure*}[htbp]
\centering
\includegraphics[width=1.0\linewidth]{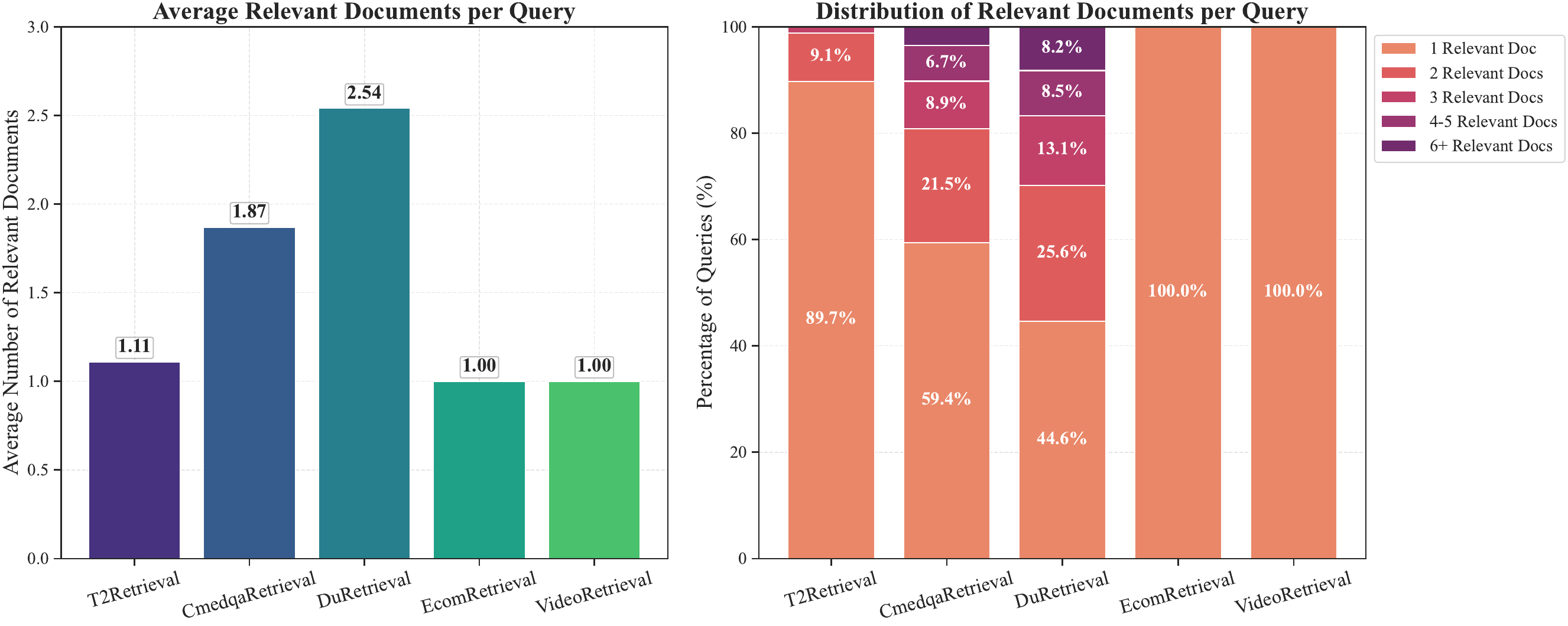}
\caption{Average number and distribution of relevant documents per query across five widely used open-source IR datasets.}
\label{fig:positives}
\end{figure*}

We analyze five open-source IR datasets to determine the average number of relevant documents per query, with results shown in Figure~\ref{fig:positives}. It is evident that even within a vast corpus, content directly relevant to a specific query is typically sparse, making it feasible to enumerate most positive samples. Furthermore, since mainstream IR datasets predominantly use binary labels (1 for positives, 0 for negatives), improving nDCG@k is equivalent to maximizing the predicted scores of a query's $\text{m} = \min(\text{k}, \text{n\_positives})$ positive documents. This objective aligns with the principles of contrastive learning but imposes two additional requirements: (1) the documents involved in the relevance comparison should come from the same corpus and be as numerous as possible, and (2) a sufficient number of positive examples should be considered simultaneously.

In CoDiEmb, the first requirement is met primarily by our sampler, which we will detail in subsection~\ref{sec:sampler_gpu}. The second is achieved through our design of an InfoNCE Loss that incorporates multiple positives and hard negatives. For a batch size of $N$, the loss is formulated as:
\begin{equation}
\label{eq:ir_loss}
    \begin{gathered}
        Z_i^{+} = \sum_{j\not=i}^N\sum_{k=1}^{K^+} e^{\cos(q_i,d^+_{jk})/\tau} \\
        Z_i^{-} = \sum_{j=1}^N\sum_{k=1}^{K^-} e^{\cos(q_i,d^-_{jk})/\tau}  \\
        \mathcal{L}_{\text{IR}} = - \mathbb{E}\left[\sum_{i=1}^N \sum_{c=1}^{K^+}\log \frac{e^{\cos(q_i,d^+_{ic})/\tau}}{e^{\cos(q_i,d^+_{ic})/\tau} + Z_i^{+} + Z_i^-} \right]
    \end{gathered}
\end{equation}

In Equation~\ref{eq:ir_loss}, $K^+$ and $K^-$ denote the number of positive and hard negative examples, drawn from $\{d^+\}_1^m$ and $\{d^-\}_1^n$, respectively. If the available samples are fewer than $K^+$ or $K^-$, we sample with replacement. By considering multiple positives against an expanded set of negatives, this contrastive objective more closely approximates the nDCG@k metric, thereby boosting performance on IR tasks.

Unlike nDCG, which is a position-aware metric that assigns greater weight to top-ranked items, the Spearman correlation coefficient $\rho$ treats each sample equally and focuses on overall ranking quality. Its formula is described in Equation~\ref{eq:spearman}, where $n$ is the number of data points, and $d_i$ is the difference in ranks between the predicted and true scores for the $i$-th pair. Spearman's coefficient ranges from -1 to 1, with higher values indicating stronger agreement between the model's output and human judgment.
\begin{equation}
\label{eq:spearman}
\rho = 1 - \frac{6\sum_{i=1}^n d_i^2}{n(n^2-1)}
\end{equation}

Training data for STS tasks often contain fine-grained annotation scores, for which coarse-grained modeling approaches like contrastive learning are often suboptimal, as they fail to fully leverage these nuances and thus face a performance ceiling. To address this, \citet{Pcc-tuning-EMNLP-2024} proposed a two-stage optimization scheme featuring a Pearson Loss that directly optimizes the model at the rank level. CoDiEmb adopts this strategy. Given a set of text pairs $\{(x^i_1, x^i_2)\}_{i=1}^n$, let the model's predicted cosine similarities be $X = \{\cos\left(f(x^i_1), f(x^i_2)\right)\}_{i=1}^n$ and the list of ground-truth similarities be $Y = \{y^i\}_{i=1}^n$. The Pearson Loss is calculated as:
\begin{equation}
\label{eq:pearson}
    \begin{gathered}
        \text{Cov}(X, Y) = \mathbb{E}\left[(X-E[X])(Y-E[Y]) \right] \\
        r = \frac{\text{Cov}(X, Y)}{\sigma_{X}\sigma_Y} = \frac{\sum_{i=1}^n (x^i - \bar{x})(y^i - \bar{y})}{\sqrt{\sum_{i=1}^n (x^i - \bar{x})^2} \sqrt{\sum_{i=1}^n (y^i - \bar{y})^2}} \\
        \mathcal{L}_{\text{pearson}} = -r + 1
    \end{gathered}
\end{equation}

While effective, Pearson correlation primarily captures linear relationships. To model more complex mappings, CoDiEmb introduces two additional list-wise losses.

KL divergence is widely used to measure the distance between a predicted distribution $Q$ and a true distribution $P$, defined as $\text{D}_{\text{KL}}(P||Q) = \sum_i^n p_i \log \frac{p_i}{q_i}$. To apply KL divergence to STS, one must first convert score distributions into probability distributions. An intuitive method is to apply the Softmax function with temperature $\tau$ to both the predicted scores $X$ and ground-truth scores $Y$ to obtain $q_i$ and $p_i$:
\begin{equation}
\label{eq:normal_kl}
    \begin{gathered}
        \hat{y}_i = \cos\left(f(x^i_1), f(x^i_2)\right)\\
        q_i = \frac{\exp(\hat{y}_i / \tau)}{\sum_{j=1}^{N} \exp(\hat{y}_j / \tau)} \\
        p_i = \frac{\exp(y_i / \tau)}{\sum_{j=1}^{N} \exp(y_j / \tau)} 
    \end{gathered}
\end{equation}

Since $p_i$ is derived from ground-truth labels and carries no gradients, optimizing KL divergence is equivalent to minimizing the cross-entropy, i.e., $\arg \min \left(\text{D}_{\text{KL}}(P||Q) \right) = \arg \min \left(-\sum_i p_i \log q_i \right)$. This process is analogous to knowledge distillation with soft labels and is logically sound. However, this approach is unstable because $p_i$ is highly sensitive to the score distribution within a batch. Consider two batches: $Y_A = [0.9, 0.88, 0.2]$ and $Y_B = [0.6, 0.2, 0.1]$. With $\tau = 0.1$, we have $P_A = \text{Softmax}(Y_A) \approx [0.5496, 0.4499 , 0.0005]$. Here, the first two samples account for 99.95\% of the total probability mass, forcing the model to expend significant effort on fitting the minute difference between scores 0.9 and 0.88, while the 0.2-scored sample receives a negligible gradient. Similarly, for batch B, $P_B = \text{Softmax}(Y_B) \approx [0.9756, 0.0179, 0.0066]$. In this case, the model is heavily incentivized to rank the first sample correctly, while the relative order of the other two is largely ignored.

This unstable weight allocation mechanism deviates from the spirit of Spearman correlation, which prioritizes rank consistency over score magnitude. We therefore propose a Normalized Rank KL-divergence Loss $\mathcal{L}_\text{RankKL}$. Instead of comparing scores, we align the predicted distribution with an ideal distribution derived from ground-truth ranks. First, we sort all samples within a batch in descending order based on their labels to obtain ranks $r_i \in [0, N-1]$, where $N$ is the batch size. In case of ties, following the definition of Spearman correlation, $r_i$ is set to the average of their ranks. These ranks are then normalized to $y'_i \in [0,1]$, aligning their scale with the predicted cosine similarities $\hat{y}$. We then define the target distribution $p_i$ as the Softmax of $y_i'$, while keeping the predicted distribution $q_i$ as before. The final loss is:
\begin{equation}
\label{eq:rank_kl}
    \begin{gathered}
        y'_i = \frac{(N - 1) - r_i}{N - 1} \\
        p_i = \frac{\exp(y'_i / \tau)}{\sum_{j=1}^{N} \exp(y'_j / \tau)} \\
        q_i = \frac{\exp(\hat{y}_i / \tau)}{\sum_{j=1}^{N} \exp(\hat{y}_j / \tau)} \\
        \mathcal{L}_\text{RankKL} = \sum_{i=1}^{N} p_i \log\left(\frac{p_i}{q_i}\right)
    \end{gathered}
\end{equation}

Compared to the original KL divergence, $\mathcal{L}_\text{RankKL}$ directly uses rank as its optimization target, making it robust to the absolute magnitudes of ground-truth scores. This allows it to provide a stable gradient throughout training, driving the predicted ranking toward the desired order.

Currently, mainstream approaches to ranking objectives often rely on penalizing inverted pairs. RankNet \citep{Rank-ICML-2005} and Cosent \citep{CoSENT-2024} are representative losses of this category. As shown in Equation~\ref{eq:rank_net}, when the ground truth is $y_i > y_j$ but the model predicts the opposite, these functions incur a loss proportional to the margin of error $\hat{y}_j - \hat{y}_i$.
\begin{equation}
\label{eq:rank_net}
    \begin{gathered}
        \mathcal{L}_{\text{RankNet}} = \sum_{i=1}^N\sum_{j=1}^N \mathbb{1}_{y_i > y_j} \log \left(1 + \exp \left(\hat{y_j} - \hat{y_i} \right )\right) \\
        \mathcal{L}_{\text{Cosent}} = \log \left( 1 + \sum \mathbb{1}_{y_i > y_j} \exp \left(\frac{\hat{y_j} - \hat{y_i}}{\tau}\right) \right)
    \end{gathered}
\end{equation}

However, while RankNet and Cosent are adaptive to some extent, they do not fully utilize the ground-truth information. Consider three samples with scores $y_i=1.0, y_j=0.5, y_k=0.0$. If the model predicts $\hat{y}_i=0.5, \hat{y}_j=0.5, \hat{y}_k=1.0$, both losses would penalize the pairs $(\hat{y}_i, \hat{y}_k)$ and $(\hat{y}_j, \hat{y}_k)$ equally. This is counter-intuitive, as the deviation for $(\hat{y}_i, \hat{y}_k)$ is clearly larger. In other words, the ground-truth scores should modulate the loss calculation itself, not merely serve as a filtering condition.

To remedy this, we adapt Preference Rank Optimization (PRO), a reinforcement learning method originally from BEQUE \citep{BEQUE-WWW-2024} for query rewriting. Similar to $\mathcal{L}_\text{RankKL}$, we first sort samples by their true scores $y_i$. For any pair $(i, j)$ in the sorted list where $i > j$ (and thus $y_i > y_j$), we define a weight $\mathcal{T}_{i}^j= \tau / (y_i - y_j)$, where $\tau$ is a temperature hyperparameter and $y_i - y_j$ is the difference in ground-truth scores. We then set $\mathcal{T}_i^i$ to $\min_{i > j}(\mathcal{T}_i^j)$, which corresponds to the pair involving sample $i$ with the largest score difference. The $\mathcal{L}_{\text{PRO}}$ is formulated as:
\begin{equation}
\label{eq:pro_loss}
\mathcal{L}_{\text{PRO}} = - \mathbb{E}\left[ \sum_{i=1}^{N-1} \log\frac{\exp(\hat{y}_i /\mathcal{T}_i^i)}{\sum_{j=i}^N\exp(\hat{y}_j /\mathcal{T}_{i}^j)} \right]
\end{equation}

$\mathcal{L}_{\text{PRO}}$ decomposes the ranking objective into $N-1$ sequential subproblems. For each anchor point $i$ in the list, we construct a classification task where the goal is to make its predicted score $\hat{y}_i$ higher than all subsequent items, with the optimization weighted by their true similarity differences.

Furthermore, inspired by curriculum learning and recognizing that list-wise objectives are harder to optimize than pair-wise ones, CoDiEmb incorporates an auxiliary InfoNCE Loss at an intermediate layer of the model:
\begin{equation}
\label{eq:mid_info}
\mathcal{L}_{\text{MidNCE}} = - \mathbb{E}\left[\sum_{i=1}^N \log \frac{\mathbb{1}_{\text{label}}  e^{\cos\left(f(x_1^i), f(x_2^i)\right)/ \tau}}{\sum_{j=1}^N  e^{\cos\left(f(x_1^i), f(x_2^j) \right)/ \tau}}  \right]
\end{equation}

Here, the indicator function $\mathbb{1}_{\text{label}}$ filters out text pairs with similarity scores below a given threshold, ensuring that the numerator of the contrastive loss consists of true positive samples. With this approach, we expect the model to first learn pair-level semantic distinctions in shallower layers before mastering more complex list-wise relationships in deeper layers. Additionally, as PLMs become deeper, applying an intermediate loss can help mitigate the vanishing gradient problem and improve optimization signal propagation \citep{DIEN-AAAI-2019}.

Finally, the total loss for STS tasks in CoDiEmb is a weighted sum of these components: $\mathcal{L}_{\text{STS}} = \alpha \mathcal{L}_{\text{Pearson}} + \beta \mathcal{L}_{\text{RankKL}} + \gamma \mathcal{L}_{\text{PRO}} + \lambda \mathcal{L}_{\text{MidNCE}}$. During training, we alternate between $\mathcal{L}_{\text{IR}}$ and $\mathcal{L}_{\text{STS}}$ to update network parameters, preventing catastrophic forgetting and achieving a robust balance across all tasks.

\subsection{Sampler and Multi-GPU Setup}
\label{sec:sampler_gpu}

\begin{figure*}[htbp]
\centering
\includegraphics[width=1.0\linewidth]{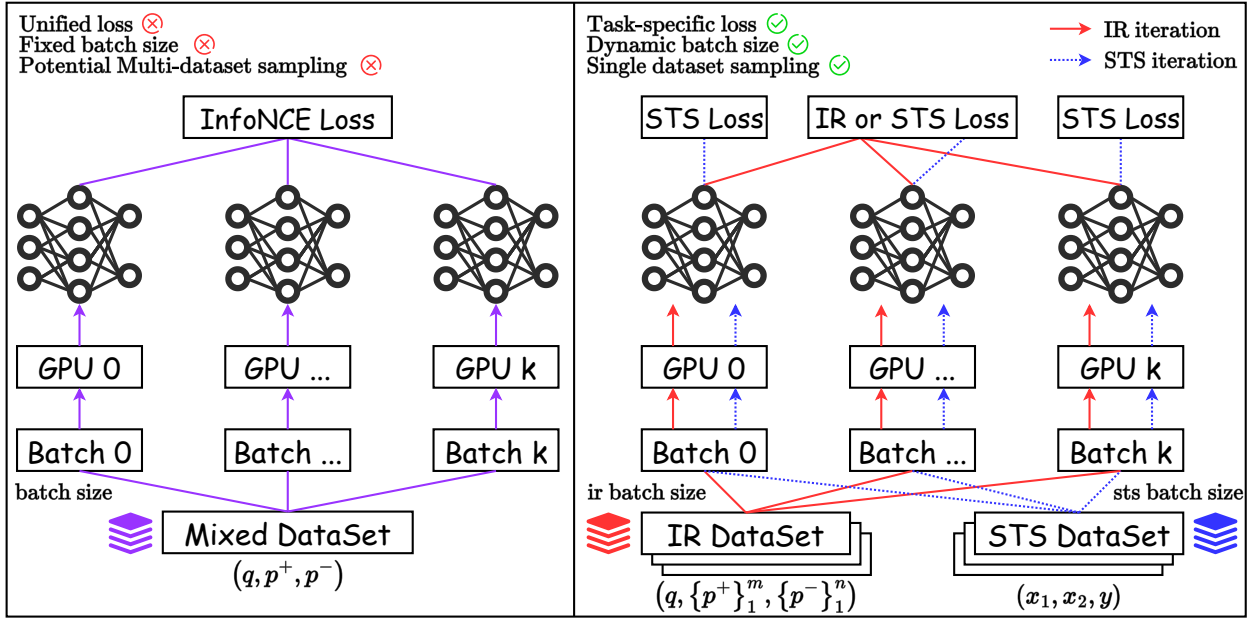}
\caption{An illustration of CoDiEmb's multi-node training framework. Compared to previous methods, we enforce strict single-dataset sampling and support distinct loss functions and batch sizes for different tasks.} 
\label{fig:sampler}
\end{figure*}

As model parameter counts and data volumes scale, distributed training has become standard practice in representation learning. Our prior analysis has highlighted that a core aspect of IR is making positive examples stand out from the entire document corpus. Thus, with appropriate learning rates and iteration counts, a model's IR performance generally benefits from larger batch sizes, a finding has been validated in several previous works \citep{Uni-Retriever-SIGKDD-2022, ESimCSE-COLING-2022, CoT-BERT-ICANN-2024}. Accordingly, as shown on the right side of Figure~\ref{fig:sampler}, CoDiEmb enables cross-device negative sampling when processing IR tasks to gather a larger pool of reference items.

However, merely increasing the number of contrastive samples is insufficient for robust performance gains; the negatives obtained from other GPUs must be meaningful. In both real-world IR applications and benchmarks, a document is ranked against others from the same corpus. Therefore, negatives drawn from the same data distribution are more challenging and informative than random documents from a global pool. Consequently, CoDiEmb implements a custom data sampler that guarantees, within a single iteration, that each device processes a non-overlapping shards of the same data file.

Conversely, for STS tasks, our empirical findings show that model convergence is not contingent on massive batch sizes. In fact, since many STS datasets use a small set of discrete integer labels (e.g., 1 to 5), an excessively large batch can lead to a high frequency of tied scores. Such a distribution can degrade the performance of rank-sensitive list-wise losses. Therefore, we disable cross-device sampling when processing STS task batches.

Furthermore, the significant disparity in typical text lengths between IR (long documents) and STS (short sentences) makes a uniform batch size inefficient, leading to unbalanced GPU utilization and difficulty in managing per-task training iterations. To resolve this, CoDiEmb's data loader supports task-specific batch size configurations, optimizing training efficiency and providing finer control over the learning process.

\subsection{Hierarchical Model Fusion}
\label{sec:model_fusion}

The practice of merging checkpoints from multiple training trajectories, often referred to as Model Soups, has been demonstrated as an effective technique for enhancing model performance in recent works like Qwen3-Embedding \citep{zhang2025qwen3} and Gemini Embedding \citep{lee2025gemini}. Given a set of $k$ fine-tuned checkpoints with parameters $\{\theta_{\text{tuned}}^1, \dots, \theta_{\text{tuned}}^k\}$, the standard Model Soups approach creates a fused model, $\theta_{\text{fused}}$, by taking a weighted average of the entire parameter sets:
\begin{equation}
\label{eq:model_soups}
\theta_{\text{fused}} = \sum_{i=1}^k w_i \theta_{\text{tuned}}^i
\end{equation}

Here, the weight $w_i$ is applied uniformly to all parameters within a given checkpoint $\theta_{\text{tuned}}^i$. This model-level strategy, however, overlooks the differential contributions of internal parameter structures in adapting to diverse tasks.

\begin{figure*}[htbp]
\centering
\includegraphics[width=1.0\linewidth]{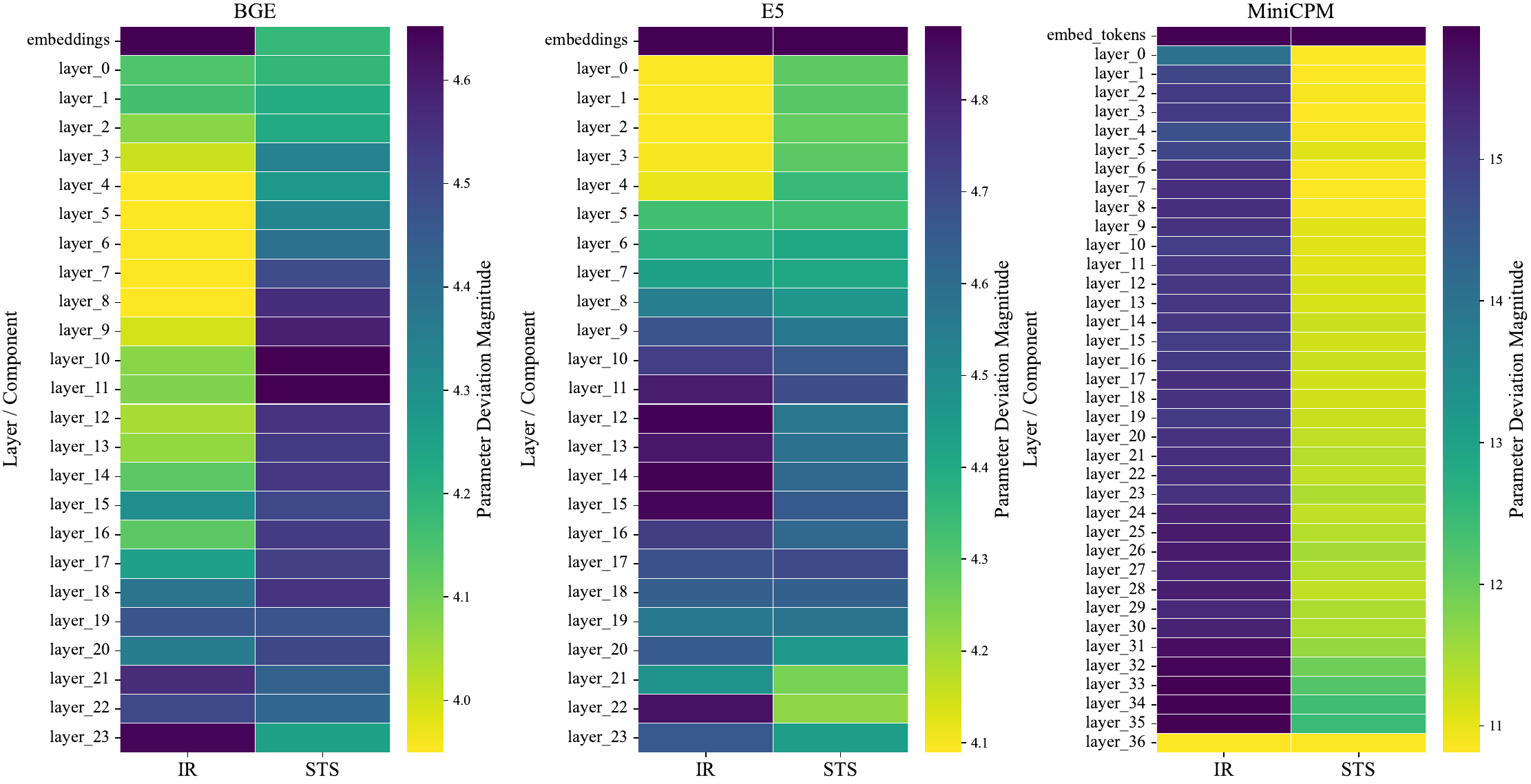}
\caption{Layer-wise parameter update magnitudes for BGE, E5, and MiniCPM models when fine-tuned only on IR and STS tasks. The color intensity represents the L2 norm of parameter deviation from the pre-trained base model. The distinct patterns for IR and STS across different layers provide empirical evidence for task-specific specialization, motivating our hierarchical fusion approach.}
\label{fig:heatmap}
\end{figure*}

To achieve a more granular fusion, we first conducted a preliminary experiment to quantify how different model components specialize. Specifically, we fine-tuned two task-specialist models using only IR and STS training data, respectively. We then measured the importance of each model layer for each task by calculating the L2 norm of its parameter deviation from the original base model. As visualized in Figure~\ref{fig:heatmap}, the heatmaps of layer-wise parameter updates across three base models reveal that different layers indeed exhibit significant variations in their adaptation to IR and STS tasks, providing empirical justification for our fine-grained fusion strategy.

Based on this insight, we propose a hierarchical fusion method. This method begins by constructing two task-biased soups. Specifically, the IR soup ($\tilde{\theta}_{\text{IR}}$) is formed by the average of models that are either well-balanced or particularly excel at the IR task. Similarly, the STS soup ($\tilde{\theta}_{\text{STS}}$) is formed by averaging models that are well-balanced or demonstrate superior performance on the STS task.  Subsequently, we use the layer-wise parameter deviation norms obtained from our preliminary experiment, denoted as $\delta_{\text{IR}, l}$ and $\delta_{\text{STS}, l}$, to calculate the fusion weights for each layer $l$ via a Softmax function with a temperature parameter $\tau$:
\begin{equation}
w_\text{IR}^{l} = \frac{e^{\delta_\text{IR}^{l}/ \tau}}{e^{\delta_\text{IR}^{l}/ \tau} + e^{\delta_\text{STS}^{l}/ \tau}},\quad w_\text{STS}^l = 1-w_\text{IR}^l
\end{equation}

Finally, we use these layer-wise weights to perform a weighted merge of the two soups, yielding the final model parameters $\theta_{\text{fused}}$:
\begin{equation}
\theta_\text{fused}^l = w_\text{IR}^l \cdot \tilde{\theta}_\text{IR}^l + w_\text{STS}^l \cdot \tilde{\theta}_\text{STS}^l
\end{equation}

\section{Experiments}
\label{sec:experiments}

This section provides empirical validations of our proposed framework, CoDiEmb. We begin in subsection~\ref{sec:implementation} by detailing our experimental setup, including evaluation benchmarks, training data, and base models. Subsequently, subsection~\ref{sec:main_performance} presents our main results and ablation studies, demonstrating the effectiveness of CoDiEmb's specialized loss functions and sampling strategy. Finally, in subsection~\ref{sec:vs_single}, we compare CoDiEmb with models trained on single tasks to fully illustrate the advantages of our joint optimization framework.

\subsection{Implementation Details}
\label{sec:implementation}

Our experiments are primarily conducted on the well-established CMTEB leaderboard \citep{CMTEB-BGE-SIGIR-2024}, which comprises 7 STS tasks and 8 IR tasks from diverse domains such as medicine, finance, and general knowledge. For training, we adopt the publicly available C-MTEB IR and STS datasets. Notably, three IR tasks—CovidRetrieval, MMarcoRetrieval, and MedicalRetrieval—do not provide dedicated training sets. Evaluations on these tasks are therefore performed in a zero-shot setting, which directly reflects the models' generalization capabilities.

To demonstrate the generality of our approach, we fine-tune three different PLMs as backbones: MiniCPM-Embedding \citep{MiniCPM-COLM-2024}, multilingual-e5-large \citep{M-E5-2024}, and bge-large-zh-v1.5 \citep{CMTEB-BGE-SIGIR-2024}. Following prior work \citep{NV-Embed-2024}, we design task-specific instructions and prepend them to all input texts. For MiniCPM and E5, we use mean pooling over the last hidden state to derive text representations, and tokens corresponding to the instructions are masked out. For BGE, we employ CLS pooling to maintain consistency with its pre-training configuration. To improve computational efficiency, we leverage DeepSpeed ZeRO-1 and enable gradient checkpointing during training.

\subsection{Main Results}
\label{sec:main_performance}
\begin{table*}[htbp]
\caption{Spearman correlation scores of different methods on the 7 STS tasks in CMTEB. The last two columns, Avg IR and Avg STS, represent the model's average performance on IR and STS, respectively. Detailed IR results are in Table~\ref{tab:ir_results}.} 
\centering
\resizebox{1.0\linewidth}{!}{
    \begin{tabular}{cccccccc|c|c}
    \toprule 
    \bf{Methods} & \bf{AFQMC} & \bf{ATEC} & \bf{BQ} & \bf{LCQMC} & \bf{PAWSX} & \bf{QBQTC} & \bf{STS-B} & \bf{Avg IR} & \bf{Avg STS} \\
    \midrule
    \multicolumn{10}{c}{\it{Implementation on} $\text{MiniCPM-Embedding}$} \\
    InfoNCE & 61.51	& 58.03	& 67.78	& 71.89	& 40.93	& 41.82	& 81.73 & \underline{74.23} & 60.53 \\
    CoSENT & 69.28 & 59.54 & 73.57 & 79.97 & 63.95 & 58.35 & 85.69 & 71.30 & 70.05 \\
    Mixed & 70.77 & 61.37 & 72.01 & 78.40 & 65.48 & 59.29 & 84.93 & 73.05 & \underline{70.32} \\
    CoDiEmb & 69.70 & 60.56 & 74.23 & 80.38 & 67.12 & 60.98 & 85.11 & \textbf{75.73} & \textbf{71.15} \\
    \midrule
    \multicolumn{10}{c}{\it{Implementation on} $\text{multilingual-e5-large}$} \\
    InfoNCE & 52.07 & 53.12 & 69.72 & 72.83 & 26.99 & 40.46 & 79.02 & \textbf{70.90} & 56.32 \\
    CoSENT & 53.18 & 53.09 & 72.19 & 80.29 & 57.53 & 53.52 & 82.50 & 65.69 & 64.61 \\
    Mixed & 58.36 & 54.91 &	72.83 &	79.99 &	63.44 &	56.87 &	81.84 & 68.61 & \underline{66.89} \\ 
    CoDiEmb & 60.81 & 56.06 & 73.08 & 80.17 & 65.41 & 57.85 & 82.20 & \underline{70.39} & \textbf{67.94} \\
    \midrule
    \multicolumn{10}{c}{\it{Implementation on} $\text{bge-large-zh-v1.5}$} \\
    InfoNCE &  54.47 &	54.34 &	68.64 &	74.16 &	34.31 &	41.12 & 79.61 & \textbf{71.73} & 58.09 \\
    CoSENT & 56.73 & 54.52 & 72.55 & 80.54 & 55.34 & 52.56 & 80.67 & 66.55 & 64.70 \\
    Mixed & 62.59 & 56.59 & 73.04 & 80.16 & 59.51 & 56.82 & 81.13 & 68.67 & \underline{67.12} \\
    CoDiEmb & 65.54 & 58.03 & 72.42 & 80.21 & 60.43 & 57.07 & 80.77 & \underline{71.14} & \textbf{67.78} \\
    \bottomrule
    \end{tabular}
}
\label{tab:sts_results}
\end{table*}

Tables~\ref{tab:sts_results} and \ref{tab:ir_results} present the detailed performance of different methods on the full set of STS and IR tasks in CMTEB, respectively. To isolate the contributions of CoDiEmb's components, we compare it against several carefully designed baselines. 

In these tables, "InfoNCE" denotes training with only the InfoNCE Loss. For STS tasks under this setting, samples with low similarity scores are filtered out via a threshold to ensure the correctness of the contrastive objective. Conversely, "CoSENT" refers to training with only the CoSENT Loss. For IR tasks under this setting, we assign a label of 1 to all positive pairs and 0 to negative pairs; this does not introduce bias, as the labels merely serve as a filter for CoSENT and do not participate directly in the loss calculation. The specific formulas for the InfoNCE and CoSENT losses used in these baselines are shown below, where the notation is consistent with previous sections. Additionally, "Mixed" in the tables refers to the adoption of a mixed-batch sampler during training. While this sampler still requires that texts within each GPU originate from the same data file, it places no such restriction across GPUs. Consequently, in a single iteration, different GPUs may process different task types, providing the model with mixed-task gradients. For a fair comparison, all methods listed in the tables are trained on the identical datasets and with the same base models.
\begin{equation}
\label{eq:infonce_cosent}
\begin{gathered}
    \mathcal{L}_{\text{InfoNCE}} = - \mathbb{E}\left[\sum_{i=1}^N \log \frac{\mathbb{1}_{\text{label}}  e^{\cos\left(f(x_1^i), f(x_2^i)\right)/ \tau}}{\sum_{j=1}^N  e^{\cos\left(f(x_1^i), f(x_2^j) \right)/ \tau}}  \right] \\
    \mathcal{L}_{\text{Cosent}} = \log \left( 1 + \sum \mathbb{1}_{y_i > y_j} \exp \left(\frac{\cos\left(f(x_1^j), f(x_2^j)\right) - \cos\left(f(x_1^i), f(x_2^i)\right)}{\tau}\right) \right) 
\end{gathered}
\end{equation}

The experimental results show that compared to using only the InfoNCE Loss, CoDiEmb achieves comparable performance on IR tasks but is substantially better on STS tasks, leading to a significantly higher overall score. This phenomenon is interpretable: when harnessing a unified contrastive learning approach, the threshold-filtered STS samples are converted into a format identical to that of IR data, effectively acting as data augmentation for the IR task. However, this marginal improvement on IR comes at the expense of a significant degradation in STS performance. For this reason, CoDiEmb avoids utilizing coarse-grained contrastive learning as the primary optimization method for STS.

Compared to using only the CoSENT Loss, CoDiEmb demonstrates markedly superior performance on both IR and STS tasks. For instance, with multilingual-e5-large as the backbone, CoDiEmb achieves gains of 4.70 on average nDCG@10 for IR and 3.33 on average Spearman correlation for STS. This highlights the inadequacy of a single pair-wise ranking loss for the distinct optimization of both tasks.

A similar trend is observed when comparing against the mixed-gradient sampler. By strictly ensuring that each GPU processes a disjoint subset of the same dataset per iteration while flexibly balancing the update frequencies of different data sources, CoDiEmb achieves robust gains across all tasks. Collectively, these ablations confirm that CoDiEmb's specialized loss functions and its single-task, multi-device sampling strategy are crucial and effective components of the framework.
\begin{table*}[htbp]
\caption{nDCG@10 scores of different methods on the 8 IR tasks in CMTEB. The last two columns, Avg IR and Avg STS, represent the model's average performance on IR and STS, respectively. Detailed STS results are in Table~\ref{tab:sts_results}.}
\centering                    
\resizebox{1.0\linewidth}{!}{
    \begin{tabular}{ccccccccc|c|c}
    \toprule 
    \bf{Methods} & \bf{Cmedqa} & \bf{Covid} & \bf{Du} & \bf{Ecom} & \bf{MMarco} & \bf{Medical} & \bf{T2} & \bf{Video} & \bf{Avg IR} & \bf{Avg STS} \\
    \midrule
    \multicolumn{11}{c}{\it{Implementation on} $\text{MiniCPM-Embedding}$} \\
    InfoNCE & 41.99	& 90.73 & 88.78 & 65.42	& 83.76 & 61.26 & 86.91 & 74.98 & \underline{74.23} & 60.53 \\
    CoSENT & 42.28 & 81.81 & 86.70 & 65.66 & 78.89 & 59.57 & 84.52 & 70.97 & 71.30 & 70.05 \\
    Mixed & 41.82 &	90.01 &	87.62 &	64.10 & 83.21 & 59.64 & 86.22 & 71.75 & 73.05 & \underline{70.32} \\
    CoDiEmb & 45.43 & 90.61 & 89.51 & 69.24 & 84.26 & 62.86 & 87.36 & 76.55 & \textbf{75.73} & \textbf{71.15} \\
    \midrule
    \multicolumn{11}{c}{\it{Implementation on} $\text{multilingual-e5-large}$} \\
    InfoNCE & 41.85 & 74.97 & 85.90 & 65.76 & 77.40 & 59.82 & 84.47 & 77.00 & \textbf{70.90} & 56.32 \\
    CoSENT & 33.49 & 70.31 & 83.82 & 62.50 & 73.05 & 51.66 & 82.54 & 68.17 & 65.69 & 64.61 \\
    Mixed & 38.53 & 75.79 & 83.02 & 63.64 & 75.75 & 55.92 & 82.12 & 74.07 & 68.61 & \underline{66.89} \\
    CoDiEmb &	40.26 &	77.13 & 85.41 & 65.16 & 77.71 & 57.32 & 84.25 & 75.87 & \underline{70.39} & \textbf{67.94} \\
    \midrule
    \multicolumn{11}{c}{\it{Implementation on} $\text{bge-large-zh-v1.5}$} \\
    InfoNCE & 45.14 & 75.86 & 88.19 & 67.33 & 76.01 & 59.51 & 84.92 & 76.86 & \textbf{71.73} & 58.09 \\
    CoSENT & 39.66 & 68.22 & 85.68 & 63.79 & 67.28 & 55.90 & 82.36 & 69.49 & 66.55 & 64.70 \\
    Mixed & 43.04 & 74.90 & 85.58 & 64.96 & 66.58 & 57.39 & 83.50 & 73.40 & 68.67 & \underline{67.12} \\
    CoDiEmb & 44.59 & 76.87 & 88.05 & 67.03 & 71.99 & 59.17 & 84.94 & 76.50 & \underline{71.14} & \textbf{67.78} \\
    \bottomrule
    \end{tabular}
}
\label{tab:ir_results}
\end{table*}

\subsection{Comparison with Single-Task Models}
\label{sec:vs_single}

To conclusively demonstrate that CoDiEmb achieves more than just a simple trade-off, we compare it against two specialist models: one trained exclusively on IR data (IR-only) and another on STS data (STS-only). The results, presented in Table~\ref{tab:vs_only}, reveal the synergistic benefits of our joint optimization framework.

Compared to the IR-only specialist, CoDiEmb incurs a negligible drop in average IR performance (approx. 1 point) but attains a massive gain in STS performance, outperforming the IR specialist by more than 16 points on average. This trade-off is highly favorable, leading to a substantially higher overall score.

More strikingly, when compared to the STS-only specialist, CoDiEmb is superior on both task types. For example, on the MiniCPM backbone, CoDiEmb outperforms the STS specialist by 13.45 points on IR and 2.32 points on STS. This demonstrates that the STS task, which is often difficult to improve, does not suffer from negative transfer but instead benefits synergistically from co-training with IR data under CoDiEmb's collaborative-distinct paradigm.
\begin{table}[htbp]
\centering
\caption{Performance comparison between CoDiEmb and models trained exclusively on single-task data. Scores are averages on IR (nDCG@10) and STS (Spearman's $\rho \times 100$) benchmarks.} 
\begin{tabular}{c|c|c|c|c}
    \toprule
    \bf PLMs & \bf Method & \bf Avg IR & \bf Avg STS & \bf Avg \\
    \midrule
    \multirow{3}{*}{MiniCPM-Embedding}
    & CoDiEmb & 75.73 & 71.15 & \textbf{73.44} \\
    & IR only & 76.10 & 49.67 & 62.89 \\
    & STS only & 62.28 & 68.83 & 65.56 \\
    \midrule
    \multirow{3}{*}{multilingual-e5-large} 
    & CoDiEmb & 70.39 & 67.94 & \textbf{69.17} \\
    & IR only & 71.34 & 48.22 & 59.78 \\
    & STS only & 48.02 & 66.37 & 57.20 \\
    \midrule
    \multirow{3}{*}{bge-large-zh-v1.5} 
    & CoDiEmb & 71.14 & 67.78 & \textbf{69.46} \\
    & IR only & 72.12 & 51.72 & 61.92 \\
    & STS only & 45.64 & 66.26 & 55.95 \\
    \bottomrule
\end{tabular}
\label{tab:vs_only}
\end{table}

\section{Analysis}
\label{sec:analysis}

To understand why CoDiEmb excels at joint optimization, this section moves beyond benchmark scores to quantitatively analyze the intrinsic geometric properties of the learned embedding space.

In representation learning, anisotropy and over-smoothing are two common critical issues that degrade the quality of embeddings. Anisotropy \citep{Anisotropy-EMNLP-2019} describes a condition where embeddings occupy a narrow cone in the vector space, leading to limited expressiveness. This phenomenon often arises from biases introduced by factors such as word frequency \citep{BERT-flow-EMNLP-2020}, capitalization \citep{CARDS-SIGIR-2022}, punctuation, and subword tokenization \citep{PromptBERT-EMNLP-2022}. Over-smoothing \citep{Over-smoothing-2022}, in contrast, occurs when the model loses the ability to distinguish between different tokens within a text, mapping them to overly similar embeddings. Both phenomena are particularly detrimental because in existing text representation methods, text embeddings are derived from token embeddings, as any bias or information loss at the token level directly compromises the quality of the final representation. Our central hypothesis is that CoDiEmb’s "Collaborative yet Distinct" architecture, with its task-specific losses and pure gradient signals, is uniquely suited to mitigate these problems. 

To test this hypothesis, we employ a suite of metrics to diagnose the health of the embedding space. Given an input sentence \(T = [t_1, t_2, \ldots, t_n]\), the model outputs a token embedding matrix $X = \{x_1, x_2, ..., x_n\} \in \mathbb{R}^{n \times d}$, We first quantify over-smoothing using Token-wise Similarity (TokSim), which calculates the average cosine similarity between all distinct token pairs \citep{SSCL-ACL-2023}. Obviously, the higher $\text{TokSim}(X)$, the more severe the over-smoothing.
\begin{equation}
    \text{TokSim}(X) = \frac{1}{n(n-1)} \sum_{i \neq j} \frac{x_i^T x_j}{\|x_i\|_2 \|x_j\|_2}
\end{equation}

Likewise, we utilize the token embedding matrix to analyze anisotropy issue from three perspectives. We compute the Rank of matrix $X$,  where a higher rank signifies richer, less redundant information. Furthermore, we perform Singular Value Decomposition (SVD) on $X$ and analyze its singular values $\sigma_1 \geq \sigma_2 \geq ... \geq \sigma_k$. Following \citep{CSE-SFP-SIGIR-2025}, we leverage the condition number and entropy of the singular values as indicators. 

The Condition Number $\kappa(X)$ is defined as the ratio of the maximum singular value to the minimum singular value. A lower value is preferred, as it indicates a more uniform distribution of singular values.
\begin{equation}
\kappa(X) = \frac{\sigma_{\text{max}}}{\sigma_{\text{min}}} = \frac{\sigma_1}{\sigma_k}
\end{equation}

The SVD Entropy $H(X)$ also measures the uniformity of the singular value distribution in the token embedding matrix $X$. To calculate this metric, we first normalize the singular value $\sigma_i$ to a probability distribution and then apply the following formula:
\begin{equation}
\begin{aligned}
p_i &= \frac{\sigma_i^2}{\sum_{j=1}^k \sigma_j^2} \\
H(X) &= -\sum_{i=1}^k p_i \log(p_i)
\end{aligned}
\end{equation}

Higher entropy indicates that more semantic dimensions contribute meaningfully to the representation, signaling a lower degree of anisotropy. 

We computed these four metrics on the seven STS test sets from C-MTEB, using BGE as the backbone model. As shown in Table~\ref{tab:analysis_metrics_en}, CoDiEmb demonstrates a consistent and stable advantage across all metrics. It achieves the lowest Token-wise Similarity, effectively mitigating over-smoothing. Concurrently, it systematically obtains a higher rank and SVD Entropy, alongside a substantially lower Condition Number, indicating a more expressive and isotropic embedding space. The improvement in the Condition Number is particularly striking. On average, CoDiEmb's value is 37\% lower than the strongest baseline (CoSENT), and on the STSB dataset, it represents a 92\% reduction compared to the InfoNCE baseline. 

We attribute this superior performance to CoDiEmb's core design. The framework synergistically combines a global "push-apart" force from the IR contrastive loss with a local "fine-sorting" pressure from the STS ranking losses. This dual-objective process, stabilized by a dynamic sampler that delivers pure, single-task gradients, prevents representational collapse and produces a geometrically superior embedding space.
\begin{table*}[htbp]
\centering
\caption{Comparison of embedding space quality metrics for different methods on the C-MTEB STS test sets. For Rank and SVD Entropy, higher is better. For Token Similarity(\%) and Condition Number, lower is better. Best results in each row is highlighted in bold.}
\label{tab:analysis_metrics_en}
\resizebox{\textwidth}{!}{%
\begin{tabular}{cccccccccc}
\toprule
\textbf{Method} & \textbf{Metric} & \textbf{ATEC} & \textbf{BQ} & \textbf{LCQMC} & \textbf{PAWSX} & \textbf{STSB} & \textbf{AFQMC} & \textbf{QBQTC} & \textbf{Avg} \\
\midrule
\multirow{4}{*}{CoDiEmb} 
& Rank & \textbf{14.97} & \textbf{12.85} & \textbf{10.63} & \textbf{38.44} & \textbf{19.80} & \textbf{14.62} & \textbf{9.47} & \textbf{17.25} \\
& Token Similarity & \textbf{67.67} & \textbf{72.92} & \textbf{70.61} & 61.68 & \textbf{65.77} & \textbf{67.82} & \textbf{70.22} & \textbf{68.10} \\
& SVD Entropy & \textbf{1.81} & \textbf{1.50} & 1.59 & \textbf{2.49} & \textbf{2.01} & \textbf{1.79} & \textbf{1.54} & \textbf{1.82} \\
& Condition Number & \textbf{7413.04} & \textbf{10847.82} & \textbf{19901.18} & \textbf{265.59} & \textbf{507.70} & \textbf{13129.53} & \textbf{15584.98} & \textbf{9664.26} \\
\midrule
\multirow{4}{*}{Mixed} 
& Rank & 14.45 & 12.58 & 10.61 & 38.42 & 19.52 & 14.33 & 9.35 & 17.04 \\
& Token Similarity & 67.83 & 74.91 & 71.86 & \textbf{61.58} & 66.54 & 67.93 & 70.46 & 68.73 \\
& SVD Entropy & 1.77 & 1.40 & 1.54 & 2.47 & 1.96 & 1.76 & 1.53 & 1.78 \\
& Condition Number & 19580.63 & 17888.89 & 20916.23 & 1007.10 & 6118.76 & 21357.64 & 18648.04 & 15073.90 \\
\midrule
\multirow{4}{*}{InfoNCE} 
& Rank & 14.52 & 12.58 & 10.61 & 38.41 & 19.54 & 14.36 & 9.35 & 17.05 \\
& Token Similarity & 73.22 & 78.79 & 70.69 & 70.67 & 77.78 & 72.41 & 72.92 & 73.93 \\
& SVD Entropy & 1.61 & 1.27 & \textbf{1.62} & 2.10 & 1.49 & 1.60 & 1.45 & 1.59 \\
& Condition Number & 21028.14 & 21571.02 & 22738.23 & 1245.10 & 7025.74 & 24150.67 & 20942.72 & 16957.37 \\
\midrule
\multirow{4}{*}{CoSENT} 
& Rank & 14.57 & 12.68 & 10.62 & 38.44 & 19.67 & 14.40 & 9.38 & 17.11 \\
& Token Similarity & 73.04 & 76.30 & 74.55 & 69.97 & 75.32 & 73.36 & 74.02 & 73.79 \\
& SVD Entropy & 1.58 & 1.36 & 1.42 & 2.07 & 1.58 & 1.56 & 1.38 & 1.56 \\
& Condition Number & 17917.04 & 16278.22 & 22119.27 & 348.73 & 3742.57 & 20770.19 & 19353.22 & 14361.32 \\
\bottomrule
\end{tabular}%
}
\end{table*}

\section{Related Work}

The development of robust text representations is a cornerstone of modern Natural Language Processing. Our work, CoDiEmb, builds upon and extends several key research areas, including foundational text embedding models, advanced loss function designs, and sophisticated model optimization strategies.

The field of text representation learning has progressed from static, context-independent word vectors, such as Word2Vec \citep{mikolov2013efficient} and GloVe \citep{pennington2014glove}, to dynamic, contextualized embeddings. A paradigm shift occurred with the advent of Pre-trained Language Models (PLMs), exemplified by BERT \citep{BERT-NAACL-2019}, which generate token representations sensitive to their surrounding context. To address the need for efficient semantic search over entire sentences, SBERT \citep{Sentence-BERT-EMNLP-2019} adapted PLMs using a Siamese network architecture. This enabled the creation of semantically meaningful sentence embeddings that could be directly compared using cosine similarity, establishing a dominant training methodology.
Following the blueprint established by SBERT, initial efforts to advance text embeddings primarily focused on scaling up encoder-only models. By training on massive, weakly-supervised text-pair datasets, models such as E5 \citep{M-E5-2024}, GTE \citep{GTE-2023}, and BGE \citep{CMTEB-BGE-SIGIR-2024} achieved remarkable zero-shot performance and, for a time, set the standard in the field. However, the field has recently undergone a significant paradigm shift, moving beyond scaled-up encoders to leverage the immense power of Large Language Models (LLMs) for representation learning. The underlying hypothesis is that the extensive parameterization and vast pre-training corpora of LLMs enable them to capture far richer and more nuanced semantic representations than their predecessors. This has given rise to a new generation of state-of-the-art embedding models that now define the forefront of this research area, with notable examples including Gecko \citep{lee2024gecko}, Jina-embeddings-v3 \citep{Jina-v3-2024}, NV-Embed \citep{NV-Embed-2024}, and Qwen3-Embedding \citep{zhang2025qwen3}.

Integral to the training of these embedding models is the choice of the loss function, which governs how the model learns to structure the embedding space. The predominant training paradigm is rooted in contrastive learning, where the objective is to pull semantically similar "positive" pairs closer together while pushing dissimilar "negative" pairs farther apart. In addition to the classic InfoNCE \citep{InfoNCE-2018} and CoSENT \citep{CoSENT-2024}, SimCSE \citep{SimCSE-EMNLP-2021} provided a landmark demonstration of its application to sentence embeddings. The key insight was not merely the use of in-batch negatives, but a surprisingly simple method for creating positive pairs: using a sentence to predict itself, with only standard dropout providing the necessary variation for the contrastive task. Furthermore, instead of using random negatives, some methods identify examples that the model finds difficult to distinguish from the positive anchor \citep{Hard-Negative-Mining-EMNLP-2020, zhan2021optimizing, suresh2021not}. Recent research has also explored alternative formulations. For instance, AnglE \citep{AnglE-Li-2023} proposed a loss function that explicitly optimizes for angular distance, better aligning the training objective with the cosine similarity metric used at inference. The careful design and application of these loss functions are paramount to achieving state-of-the-art results.

Model merging, the process of integrating parameters from multiple specialized models into a single, more capable artifact, has become an effective strategy for improving performance \citep{matena2022merging}. This general approach has been successfully applied to mitigate catastrophic forgetting in continual learning and to facilitate knowledge transfer in multi-task learning \citep{yu2024language}. The Model Soups \citep{Model-Soups-PMLR-2022} technique challenges the conventional practice of selecting only the single best model from a hyperparameter sweep, proposing instead to average the weights of multiple well-performing fine-tuned models. The success of this weight-averaging philosophy has inspired further research into combining knowledge from different training trajectories, such as Average Merging \citep{yang2024model} and  Fisher Merging \citep{matena2022merging}. Inspired by these strategies, we merge multiple checkpoints saved during the fine-tuning process using spherical linear interpolation (slerp). This technique aims to leverage the knowledge gained at different stages of training to produce a final model with enhanced robustness and generalization.

\section{Conclusion}

In this paper, we introduce CoDiEmb, a novel framework designed to collaboratively yet distinctly optimize text representations for the fundamental tasks of Information Retrieval (IR) and Semantic Textual Similarity (STS). CoDiEmb is built upon a suite of innovations, including a unified data format, task-differentiated loss functions, a specialized data sampler, and a hierarchical model fusion strategy. Through extensive experiments, CoDiEmb has demonstrated significant performance improvements across a diverse set of IR and STS benchmarks spanning domains such as healthcare, finance, and encyclopedias. The success of CoDiEmb suggests that the pursuit of universal embedding models should transcend conventional multi-stage contrastive learning. Instead, a more promising direction lies in developing a unified framework that explicitly leverages task-specific characteristics to attain a synergistic equilibrium. Future work will focus on extending CoDiEmb and exploring its generalization to a broader range of tasks.





\setcitestyle{numbers,square}
\setcitestyle{square,numbers,comma}
\bibliography{youtu_bib}

\begin{thebibliography}{57}
\providecommand{\natexlab}[1]{#1}
\providecommand{\url}[1]{\texttt{#1}}
\expandafter\ifx\csname urlstyle\endcsname\relax
  \providecommand{\doi}[1]{doi: #1}\else
  \providecommand{\doi}{doi: \begingroup \urlstyle{rm}\Url}\fi

\bibitem[Muennighoff et~al.(2024)Muennighoff, Hongjin, Wang, Yang, Wei, Yu, Singh, and Kiela]{GRIT-ICLR-2024}
Niklas Muennighoff, SU~Hongjin, Liang Wang, Nan Yang, Furu Wei, Tao Yu, Amanpreet Singh, and Douwe Kiela.
\newblock Generative representational instruction tuning.
\newblock In \emph{The Thirteenth International Conference on Learning Representations}, 2024.

\bibitem[Gao et~al.(2021)Gao, Yao, and Chen]{SimCSE-EMNLP-2021}
Tianyu Gao, Xingcheng Yao, and Danqi Chen.
\newblock {S}im{CSE}: Simple contrastive learning of sentence embeddings.
\newblock In \emph{Proceedings of the 2021 Conference on Empirical Methods in Natural Language Processing}, pages 6894--6910, Online and Punta Cana, Dominican Republic, November 2021. Association for Computational Linguistics.
\newblock \doi{10.18653/v1/2021.emnlp-main.552}.
\newblock URL \url{https://aclanthology.org/2021.emnlp-main.552/}.

\bibitem[Sheng et~al.(2024)Sheng, Yang, Gong, Wang, Chan, Zhang, Cheng, Zhu, Ge, Zhu, et~al.]{SimTier-CIKM-2024}
Xiang-Rong Sheng, Feifan Yang, Litong Gong, Biao Wang, Zhangming Chan, Yujing Zhang, Yueyao Cheng, Yong-Nan Zhu, Tiezheng Ge, Han Zhu, et~al.
\newblock Enhancing taobao display advertising with multimodal representations: Challenges, approaches and insights.
\newblock In \emph{Proceedings of the 33rd ACM International Conference on Information and Knowledge Management}, pages 4858--4865, 2024.

\bibitem[Sun et~al.(2025)Sun, Zhong, Zhou, and Han]{DynamicRAG-2025}
Jiashuo Sun, Xianrui Zhong, Sizhe Zhou, and Jiawei Han.
\newblock Dynamicrag: Leveraging outputs of large language model as feedback for dynamic reranking in retrieval-augmented generation.
\newblock \emph{arXiv preprint arXiv:2505.07233}, 2025.

\bibitem[Xiao et~al.(2024{\natexlab{a}})Xiao, Liu, Zhang, Muennighoff, Lian, and Nie]{CMTEB-BGE-SIGIR-2024}
Shitao Xiao, Zheng Liu, Peitian Zhang, Niklas Muennighoff, Defu Lian, and Jian-Yun Nie.
\newblock C-pack: Packed resources for general chinese embeddings.
\newblock In \emph{Proceedings of the 47th International ACM SIGIR Conference on Research and Development in Information Retrieval}, SIGIR '24, page 641–649, New York, NY, USA, 2024{\natexlab{a}}. Association for Computing Machinery.
\newblock ISBN 9798400704314.
\newblock \doi{10.1145/3626772.3657878}.
\newblock URL \url{https://doi.org/10.1145/3626772.3657878}.

\bibitem[Lee et~al.(2024{\natexlab{a}})Lee, Roy, Xu, Raiman, Shoeybi, Catanzaro, and Ping]{NV-Embed-2024}
Chankyu Lee, Rajarshi Roy, Mengyao Xu, Jonathan Raiman, Mohammad Shoeybi, Bryan Catanzaro, and Wei Ping.
\newblock Nv-embed: Improved techniques for training llms as generalist embedding models.
\newblock \emph{arXiv preprint arXiv:2405.17428}, 2024{\natexlab{a}}.

\bibitem[Deshpande et~al.(2023)Deshpande, Jimenez, Chen, Murahari, Graf, Rajpurohit, Kalyan, Chen, and Narasimhan]{C-STS-EMNLP-2023}
Ameet Deshpande, Carlos Jimenez, Howard Chen, Vishvak Murahari, Victoria Graf, Tanmay Rajpurohit, Ashwin Kalyan, Danqi Chen, and Karthik Narasimhan.
\newblock {C}-{STS}: Conditional semantic textual similarity.
\newblock In \emph{Proceedings of the 2023 Conference on Empirical Methods in Natural Language Processing}, pages 5669--5690, Singapore, December 2023. Association for Computational Linguistics.
\newblock \doi{10.18653/v1/2023.emnlp-main.345}.
\newblock URL \url{https://aclanthology.org/2023.emnlp-main.345/}.

\bibitem[Zar(2005)]{Spearman-2005}
Jerrold~H Zar.
\newblock Spearman rank correlation.
\newblock \emph{Encyclopedia of biostatistics}, 7, 2005.

\bibitem[Wang et~al.(2013)Wang, Wang, Li, He, and Liu]{nDCG-PMLR-2013}
Yining Wang, Liwei Wang, Yuanzhi Li, Di~He, and Tie-Yan Liu.
\newblock A theoretical analysis of ndcg type ranking measures.
\newblock In \emph{Proceedings of the 26th Annual Conference on Learning Theory}, volume~30 of \emph{Proceedings of Machine Learning Research}, pages 25--54, Princeton, NJ, USA, 12--14 Jun 2013. PMLR.
\newblock URL \url{https://proceedings.mlr.press/v30/Wang13.html}.

\bibitem[Oord et~al.(2018)Oord, Li, and Vinyals]{InfoNCE-2018}
Aaron van~den Oord, Yazhe Li, and Oriol Vinyals.
\newblock Representation learning with contrastive predictive coding.
\newblock \emph{arXiv preprint arXiv:1807.03748}, 2018.

\bibitem[Huang et~al.(2024)Huang, Peng, Zou, Liu, Li, Liu, Wu, Su, and Yu]{CoSENT-2024}
Xiang Huang, Hao Peng, Dongcheng Zou, Zhiwei Liu, Jianxin Li, Kay Liu, Jia Wu, Jianlin Su, and Philip~S Yu.
\newblock Cosent: consistent sentence embedding via similarity ranking.
\newblock \emph{IEEE/ACM Transactions on Audio, Speech, and Language Processing}, 32:\penalty0 2800--2813, 2024.

\bibitem[Asai et~al.(2022)Asai, Schick, Lewis, Chen, Izacard, Riedel, Hajishirzi, and Yih]{Task-Instruction-2022}
Akari Asai, Timo Schick, Patrick Lewis, Xilun Chen, Gautier Izacard, Sebastian Riedel, Hannaneh Hajishirzi, and Wen-tau Yih.
\newblock Task-aware retrieval with instructions.
\newblock \emph{arXiv preprint arXiv:2211.09260}, 2022.

\bibitem[Sturua et~al.(2024)Sturua, Mohr, Akram, G{\"u}nther, Wang, Krimmel, Wang, Mastrapas, Koukounas, Wang, et~al.]{Jina-v3-2024}
Saba Sturua, Isabelle Mohr, Mohammad~Kalim Akram, Michael G{\"u}nther, Bo~Wang, Markus Krimmel, Feng Wang, Georgios Mastrapas, Andreas Koukounas, Nan Wang, et~al.
\newblock jina-embeddings-v3: Multilingual embeddings with task lora.
\newblock \emph{arXiv preprint arXiv:2409.10173}, 2024.

\bibitem[Zhang and Li(2024{\natexlab{a}})]{Pcc-tuning-EMNLP-2024}
Bowen Zhang and Chunping Li.
\newblock Pcc-tuning: Breaking the contrastive learning ceiling in semantic textual similarity.
\newblock In \emph{Proceedings of the 2024 Conference on Empirical Methods in Natural Language Processing}, pages 14290--14302, Miami, Florida, USA, November 2024{\natexlab{a}}. Association for Computational Linguistics.
\newblock \doi{10.18653/v1/2024.emnlp-main.791}.
\newblock URL \url{https://aclanthology.org/2024.emnlp-main.791/}.

\bibitem[Zhang and Li(2024{\natexlab{b}})]{STS-Reg-EMNLP-2024}
Bowen Zhang and Chunping Li.
\newblock Advancing semantic textual similarity modeling: A regression framework with translated {R}e{LU} and smooth k2 loss.
\newblock In \emph{Proceedings of the 2024 Conference on Empirical Methods in Natural Language Processing}, pages 11882--11893, Miami, Florida, USA, November 2024{\natexlab{b}}. Association for Computational Linguistics.
\newblock \doi{10.18653/v1/2024.emnlp-main.663}.
\newblock URL \url{https://aclanthology.org/2024.emnlp-main.663/}.

\bibitem[Peng et~al.(2024)Peng, Li, Jiang, Wang, Ou, Zeng, Xu, Xu, and Chen]{BEQUE-WWW-2024}
Wenjun Peng, Guiyang Li, Yue Jiang, Zilong Wang, Dan Ou, Xiaoyi Zeng, Derong Xu, Tong Xu, and Enhong Chen.
\newblock Large language model based long-tail query rewriting in taobao search.
\newblock In \emph{Companion Proceedings of the ACM Web Conference 2024}, WWW '24, page 20–28, New York, NY, USA, 2024. Association for Computing Machinery.
\newblock ISBN 9798400701726.
\newblock \doi{10.1145/3589335.3648298}.
\newblock URL \url{https://doi.org/10.1145/3589335.3648298}.

\bibitem[Wortsman et~al.(2022)Wortsman, Ilharco, Gadre, Roelofs, Gontijo-Lopes, Morcos, Namkoong, Farhadi, Carmon, Kornblith, et~al.]{Model-Soups-PMLR-2022}
Mitchell Wortsman, Gabriel Ilharco, Samir~Ya Gadre, Rebecca Roelofs, Raphael Gontijo-Lopes, Ari~S Morcos, Hongseok Namkoong, Ali Farhadi, Yair Carmon, Simon Kornblith, et~al.
\newblock Model soups: averaging weights of multiple fine-tuned models improves accuracy without increasing inference time.
\newblock In \emph{International conference on machine learning}, pages 23965--23998. PMLR, 2022.

\bibitem[Hu et~al.(2024)Hu, Tu, Han, He, Cui, Long, Zheng, Fang, Huang, Zhao, et~al.]{MiniCPM-COLM-2024}
Shengding Hu, Yuge Tu, Xu~Han, Chaoqun He, Ganqu Cui, Xiang Long, Zhi Zheng, Yewei Fang, Yuxiang Huang, Weilin Zhao, et~al.
\newblock Minicpm: Unveiling the potential of small language models with scalable training strategies.
\newblock \emph{arXiv preprint arXiv:2404.06395}, 2024.

\bibitem[Wang et~al.(2024)Wang, Yang, Huang, Yang, Majumder, and Wei]{M-E5-2024}
Liang Wang, Nan Yang, Xiaolong Huang, Linjun Yang, Rangan Majumder, and Furu Wei.
\newblock Multilingual e5 text embeddings: A technical report.
\newblock \emph{arXiv preprint arXiv:2402.05672}, 2024.

\bibitem[Ethayarajh(2019)]{Anisotropy-EMNLP-2019}
Kawin Ethayarajh.
\newblock How contextual are contextualized word representations? {C}omparing the geometry of {BERT}, {ELM}o, and {GPT}-2 embeddings.
\newblock In \emph{Proceedings of the 2019 Conference on Empirical Methods in Natural Language Processing and the 9th International Joint Conference on Natural Language Processing (EMNLP-IJCNLP)}, pages 55--65, Hong Kong, China, November 2019. Association for Computational Linguistics.
\newblock \doi{10.18653/v1/D19-1006}.

\bibitem[Shi et~al.(2022)Shi, Gao, Xu, Liang, Li, Kong, Lee, and Kwok]{Over-smoothing-2022}
Han Shi, Jiahui Gao, Hang Xu, Xiaodan Liang, Zhenguo Li, Lingpeng Kong, Stephen Lee, and James~T Kwok.
\newblock Revisiting over-smoothing in bert from the perspective of graph.
\newblock \emph{arXiv preprint arXiv:2202.08625}, 2022.

\bibitem[Zhan et~al.(2021)Zhan, Mao, Liu, Guo, Zhang, and Ma]{zhan2021optimizing}
Jingtao Zhan, Jiaxin Mao, Yiqun Liu, Jiafeng Guo, Min Zhang, and Shaoping Ma.
\newblock Optimizing dense retrieval model training with hard negatives.
\newblock In \emph{Proceedings of the 44th international ACM SIGIR conference on research and development in information retrieval}, pages 1503--1512, 2021.

\bibitem[Zhou et~al.(2022)Zhou, Gong, Liu, Zhao, Shen, Dong, Lu, Majumder, Wen, and Duan]{SimANS-EMNLP-2022}
Kun Zhou, Yeyun Gong, Xiao Liu, Wayne~Xin Zhao, Yelong Shen, Anlei Dong, Jingwen Lu, Rangan Majumder, Ji-rong Wen, and Nan Duan.
\newblock {S}im{ANS}: Simple ambiguous negatives sampling for dense text retrieval.
\newblock In \emph{Proceedings of the 2022 Conference on Empirical Methods in Natural Language Processing: Industry Track}, pages 548--559, Abu Dhabi, UAE, December 2022. Association for Computational Linguistics.
\newblock \doi{10.18653/v1/2022.emnlp-industry.56}.
\newblock URL \url{https://aclanthology.org/2022.emnlp-industry.56/}.

\bibitem[Reimers and Gurevych(2019)]{Sentence-BERT-EMNLP-2019}
Nils Reimers and Iryna Gurevych.
\newblock Sentence-{BERT}: Sentence embeddings using {S}iamese {BERT}-networks.
\newblock In \emph{Proceedings of the 2019 Conference on Empirical Methods in Natural Language Processing and the 9th International Joint Conference on Natural Language Processing (EMNLP-IJCNLP)}, pages 3982--3992, Hong Kong, China, November 2019. Association for Computational Linguistics.
\newblock \doi{10.18653/v1/D19-1410}.
\newblock URL \url{https://aclanthology.org/D19-1410/}.

\bibitem[Burges et~al.(2005)Burges, Shaked, Renshaw, Lazier, Deeds, Hamilton, and Hullender]{Rank-ICML-2005}
Chris Burges, Tal Shaked, Erin Renshaw, Ari Lazier, Matt Deeds, Nicole Hamilton, and Greg Hullender.
\newblock Learning to rank using gradient descent.
\newblock In \emph{Proceedings of the 22nd International Conference on Machine Learning}, ICML '05, page 89–96, New York, NY, USA, 2005. Association for Computing Machinery.
\newblock ISBN 1595931805.
\newblock \doi{10.1145/1102351.1102363}.
\newblock URL \url{https://doi.org/10.1145/1102351.1102363}.

\bibitem[Zhou et~al.(2019)Zhou, Mou, Fan, Pi, Bian, Zhou, Zhu, and Gai]{DIEN-AAAI-2019}
Guorui Zhou, Na~Mou, Ying Fan, Qi~Pi, Weijie Bian, Chang Zhou, Xiaoqiang Zhu, and Kun Gai.
\newblock Deep interest evolution network for click-through rate prediction.
\newblock In \emph{Proceedings of the AAAI conference on artificial intelligence}, AAAI'19/IAAI'19/EAAI'19. AAAI Press, 2019.
\newblock ISBN 978-1-57735-809-1.
\newblock \doi{10.1609/aaai.v33i01.33015941}.
\newblock URL \url{https://doi.org/10.1609/aaai.v33i01.33015941}.

\bibitem[Zhang et~al.(2022)Zhang, Liu, Han, Xiao, Zheng, Shao, Sun, Zhu, Srinivasan, Deng, Zhang, and Xie]{Uni-Retriever-SIGKDD-2022}
Jianjin Zhang, Zheng Liu, Weihao Han, Shitao Xiao, Ruicheng Zheng, Yingxia Shao, Hao Sun, Hanqing Zhu, Premkumar Srinivasan, Weiwei Deng, Qi~Zhang, and Xing Xie.
\newblock Uni-retriever: Towards learning the unified embedding based retriever in bing sponsored search.
\newblock In \emph{Proceedings of the 28th ACM SIGKDD Conference on Knowledge Discovery and Data Mining}, KDD '22, page 4493–4501, New York, NY, USA, 2022. Association for Computing Machinery.
\newblock ISBN 9781450393850.
\newblock \doi{10.1145/3534678.3539212}.
\newblock URL \url{https://doi.org/10.1145/3534678.3539212}.

\bibitem[Wu et~al.(2022)Wu, Gao, Zang, Han, Wang, and Hu]{ESimCSE-COLING-2022}
Xing Wu, Chaochen Gao, Liangjun Zang, Jizhong Han, Zhongyuan Wang, and Songlin Hu.
\newblock {ES}im{CSE}: Enhanced sample building method for contrastive learning of unsupervised sentence embedding.
\newblock In \emph{Proceedings of the 29th International Conference on Computational Linguistics}, pages 3898--3907, Gyeongju, Republic of Korea, October 2022. International Committee on Computational Linguistics.
\newblock URL \url{https://aclanthology.org/2022.coling-1.342/}.

\bibitem[Zhang et~al.(2024)Zhang, Chang, and Li]{CoT-BERT-ICANN-2024}
Bowen Zhang, Kehua Chang, and Chunping Li.
\newblock Cot-bert: Enhancing unsupervised sentence representation through chain-of-thought.
\newblock In \emph{Artificial Neural Networks and Machine Learning – ICANN 2024: 33rd International Conference on Artificial Neural Networks, Lugano, Switzerland, September 17–20, 2024, Proceedings, Part VII}, page 148–163, Berlin, Heidelberg, 2024. Springer-Verlag.
\newblock ISBN 978-3-031-72349-0.
\newblock \doi{10.1007/978-3-031-72350-6_10}.
\newblock URL \url{https://doi.org/10.1007/978-3-031-72350-6_10}.

\bibitem[Zhang et~al.(2025{\natexlab{a}})Zhang, Li, Long, Zhang, Lin, Yang, Xie, Yang, Liu, Lin, et~al.]{zhang2025qwen3}
Yanzhao Zhang, Mingxin Li, Dingkun Long, Xin Zhang, Huan Lin, Baosong Yang, Pengjun Xie, An~Yang, Dayiheng Liu, Junyang Lin, et~al.
\newblock Qwen3 embedding: Advancing text embedding and reranking through foundation models.
\newblock \emph{arXiv preprint arXiv:2506.05176}, 2025{\natexlab{a}}.

\bibitem[Lee et~al.(2025)Lee, Chen, Dua, Cer, Shanbhogue, Naim, {\'A}brego, Li, Chen, Vera, et~al.]{lee2025gemini}
Jinhyuk Lee, Feiyang Chen, Sahil Dua, Daniel Cer, Madhuri Shanbhogue, Iftekhar Naim, Gustavo~Hern{\'a}ndez {\'A}brego, Zhe Li, Kaifeng Chen, Henrique~Schechter Vera, et~al.
\newblock Gemini embedding: Generalizable embeddings from gemini.
\newblock \emph{arXiv preprint arXiv:2503.07891}, 2025.

\bibitem[Li et~al.(2020)Li, Zhou, He, Wang, Yang, and Li]{BERT-flow-EMNLP-2020}
Bohan Li, Hao Zhou, Junxian He, Mingxuan Wang, Yiming Yang, and Lei Li.
\newblock On the sentence embeddings from pre-trained language models.
\newblock In \emph{Proceedings of the 2020 Conference on Empirical Methods in Natural Language Processing (EMNLP)}, pages 9119--9130, Online, November 2020. Association for Computational Linguistics.
\newblock \doi{10.18653/v1/2020.emnlp-main.733}.

\bibitem[Wang et~al.(2022)Wang, Ge, Zhang, and Yang]{CARDS-SIGIR-2022}
Wei Wang, Liangzhu Ge, Jingqiao Zhang, and Cheng Yang.
\newblock Improving contrastive learning of sentence embeddings with case-augmented positives and retrieved negatives.
\newblock In \emph{Proceedings of the 45th International ACM SIGIR Conference on Research and Development in Information Retrieval}, SIGIR '22, page 2159–2165, New York, NY, USA, 2022. Association for Computing Machinery.
\newblock ISBN 9781450387323.
\newblock \doi{10.1145/3477495.3531823}.

\bibitem[Jiang et~al.(2022)Jiang, Jiao, Huang, Zhang, Wang, Zhuang, Wei, Huang, Deng, and Zhang]{PromptBERT-EMNLP-2022}
Ting Jiang, Jian Jiao, Shaohan Huang, Zihan Zhang, Deqing Wang, Fuzhen Zhuang, Furu Wei, Haizhen Huang, Denvy Deng, and Qi~Zhang.
\newblock {P}rompt{BERT}: Improving {BERT} sentence embeddings with prompts.
\newblock In \emph{Proceedings of the 2022 Conference on Empirical Methods in Natural Language Processing}, pages 8826--8837, Abu Dhabi, United Arab Emirates, December 2022. Association for Computational Linguistics.
\newblock \doi{10.18653/v1/2022.emnlp-main.603}.

\bibitem[Chen et~al.(2023)Chen, Shou, Pei, Gong, Cao, Chang, Li, and Jiang]{SSCL-ACL-2023}
Nuo Chen, Linjun Shou, Jian Pei, Ming Gong, Bowen Cao, Jianhui Chang, Jia Li, and Daxin Jiang.
\newblock Alleviating over-smoothing for unsupervised sentence representation.
\newblock In \emph{Proceedings of the 61st Annual Meeting of the Association for Computational Linguistics (Volume 1: Long Papers)}, pages 3552--3566, Toronto, Canada, July 2023. Association for Computational Linguistics.
\newblock \doi{10.18653/v1/2023.acl-long.197}.

\bibitem[Zhang et~al.(2025{\natexlab{b}})Zhang, Song, and Li]{CSE-SFP-SIGIR-2025}
Bowen Zhang, Zixin Song, and Chunping Li.
\newblock Cse-sfp: Enabling unsupervised sentence representation learning via a single forward pass.
\newblock In \emph{Proceedings of the 48th International ACM SIGIR Conference on Research and Development in Information Retrieval}, pages 1402--1412, 2025{\natexlab{b}}.

\bibitem[Mikolov et~al.(2013)Mikolov, Chen, Corrado, and Dean]{mikolov2013efficient}
Tomas Mikolov, Kai Chen, Greg Corrado, and Jeffrey Dean.
\newblock Efficient estimation of word representations in vector space.
\newblock \emph{arXiv preprint arXiv:1301.3781}, 2013.

\bibitem[Pennington et~al.(2014)Pennington, Socher, and Manning]{pennington2014glove}
Jeffrey Pennington, Richard Socher, and Christopher~D Manning.
\newblock Glove: Global vectors for word representation.
\newblock In \emph{Proceedings of the 2014 conference on empirical methods in natural language processing (EMNLP)}, pages 1532--1543, 2014.

\bibitem[Devlin et~al.(2019)Devlin, Chang, Lee, and Toutanova]{BERT-NAACL-2019}
Jacob Devlin, Ming-Wei Chang, Kenton Lee, and Kristina Toutanova.
\newblock {BERT}: Pre-training of deep bidirectional transformers for language understanding.
\newblock In \emph{Proceedings of the 2019 Conference of the North {A}merican Chapter of the Association for Computational Linguistics: Human Language Technologies, Volume 1 (Long and Short Papers)}, pages 4171--4186, Minneapolis, Minnesota, June 2019. Association for Computational Linguistics.
\newblock \doi{10.18653/v1/N19-1423}.
\newblock URL \url{https://aclanthology.org/N19-1423/}.

\bibitem[Li et~al.(2023)Li, Zhang, Zhang, Long, Xie, and Zhang]{GTE-2023}
Zehan Li, Xin Zhang, Yanzhao Zhang, Dingkun Long, Pengjun Xie, and Meishan Zhang.
\newblock Towards general text embeddings with multi-stage contrastive learning.
\newblock \emph{arXiv preprint arXiv:2308.03281}, 2023.

\bibitem[Lee et~al.(2024{\natexlab{b}})Lee, Dai, Ren, Chen, Cer, Cole, Hui, Boratko, Kapadia, Ding, et~al.]{lee2024gecko}
Jinhyuk Lee, Zhuyun Dai, Xiaoqi Ren, Blair Chen, Daniel Cer, Jeremy~R Cole, Kai Hui, Michael Boratko, Rajvi Kapadia, Wen Ding, et~al.
\newblock Gecko: Versatile text embeddings distilled from large language models.
\newblock \emph{arXiv preprint arXiv:2403.20327}, 2024{\natexlab{b}}.

\bibitem[Karpukhin et~al.(2020)Karpukhin, Oguz, Min, Lewis, Wu, Edunov, Chen, and Yih]{Hard-Negative-Mining-EMNLP-2020}
Vladimir Karpukhin, Barlas Oguz, Sewon Min, Patrick Lewis, Ledell Wu, Sergey Edunov, Danqi Chen, and Wen-tau Yih.
\newblock Dense passage retrieval for open-domain question answering.
\newblock In \emph{Proceedings of the 2020 Conference on Empirical Methods in Natural Language Processing (EMNLP)}, pages 6769--6781, Online, November 2020. Association for Computational Linguistics.
\newblock \doi{10.18653/v1/2020.emnlp-main.550}.
\newblock URL \url{https://aclanthology.org/2020.emnlp-main.550/}.

\bibitem[Suresh and Ong(2021)]{suresh2021not}
Varsha Suresh and Desmond Ong.
\newblock Not all negatives are equal: Label-aware contrastive loss for fine-grained text classification.
\newblock In \emph{Proceedings of the 2021 Conference on Empirical Methods in Natural Language Processing}, pages 4381--4394, 2021.

\bibitem[Li and Li(2023)]{AnglE-Li-2023}
Xianming Li and Jing Li.
\newblock Angle-optimized text embeddings.
\newblock \emph{arXiv preprint arXiv:2309.12871}, 2023.

\bibitem[Matena and Raffel(2022)]{matena2022merging}
Michael~S Matena and Colin~A Raffel.
\newblock Merging models with fisher-weighted averaging.
\newblock \emph{Advances in Neural Information Processing Systems}, 35:\penalty0 17703--17716, 2022.

\bibitem[Yu et~al.(2024)Yu, Yu, Yu, Huang, and Li]{yu2024language}
Le~Yu, Bowen Yu, Haiyang Yu, Fei Huang, and Yongbin Li.
\newblock Language models are super mario: Absorbing abilities from homologous models as a free lunch.
\newblock In \emph{Forty-first International Conference on Machine Learning}, 2024.

\bibitem[Yang et~al.(2024)Yang, Shen, Guo, Wang, Cao, Zhang, and Tao]{yang2024model}
Enneng Yang, Li~Shen, Guibing Guo, Xingwei Wang, Xiaochun Cao, Jie Zhang, and Dacheng Tao.
\newblock Model merging in llms, mllms, and beyond: Methods, theories, applications and opportunities.
\newblock \emph{arXiv preprint arXiv:2408.07666}, 2024.

\bibitem[Qiu et~al.(2022)Qiu, Li, Qu, Chen, She, Liu, Wu, and Wang]{qiu2022dureader_retrieval}
Yifu Qiu, Hongyu Li, Yingqi Qu, Ying Chen, Qiaoqiao She, Jing Liu, Hua Wu, and Haifeng Wang.
\newblock Dureader\_retrieval: A large-scale chinese benchmark for passage retrieval from web search engine.
\newblock \emph{arXiv preprint arXiv:2203.10232}, 2022.

\bibitem[Bonifacio et~al.(2021)Bonifacio, Jeronymo, Abonizio, Campiotti, Fadaee, Lotufo, and Nogueira]{bonifacio2021mmarco}
Luiz Bonifacio, Vitor Jeronymo, Hugo~Queiroz Abonizio, Israel Campiotti, Marzieh Fadaee, Roberto Lotufo, and Rodrigo Nogueira.
\newblock mmarco: A multilingual version of the ms marco passage ranking dataset.
\newblock \emph{arXiv preprint arXiv:2108.13897}, 2021.

\bibitem[Xie et~al.(2023)Xie, Dong, Wang, Lv, Yao, Gan, Wu, Li, Li, Liu, et~al.]{xie2023t2ranking}
Xiaohui Xie, Qian Dong, Bingning Wang, Feiyang Lv, Ting Yao, Weinan Gan, Zhijing Wu, Xiangsheng Li, Haitao Li, Yiqun Liu, et~al.
\newblock T2ranking: A large-scale chinese benchmark for passage ranking.
\newblock In \emph{Proceedings of the 46th International ACM SIGIR Conference on Research and Development in Information Retrieval}, pages 2681--2690, 2023.

\bibitem[Long et~al.(2022)Long, Gao, Zou, Xu, Xie, Guo, Xu, Jiang, Xing, and Yang]{long2022multi}
Dingkun Long, Qiong Gao, Kuan Zou, Guangwei Xu, Pengjun Xie, Ruijie Guo, Jian Xu, Guanjun Jiang, Luxi Xing, and Ping Yang.
\newblock Multi-cpr: A multi domain chinese dataset for passage retrieval.
\newblock In \emph{Proceedings of the 45th International ACM SIGIR Conference on Research and Development in Information Retrieval}, pages 3046--3056, 2022.

\bibitem[Xu et~al.(2020)Xu, Hu, Zhang, Li, Cao, Li, Xu, Sun, Yu, Yu, et~al.]{xu2020clue}
Liang Xu, Hai Hu, Xuanwei Zhang, Lu~Li, Chenjie Cao, Yudong Li, Yechen Xu, Kai Sun, Dian Yu, Cong Yu, et~al.
\newblock Clue: A chinese language understanding evaluation benchmark.
\newblock \emph{arXiv preprint arXiv:2004.05986}, 2020.

\bibitem[Xiao et~al.(2024{\natexlab{b}})Xiao, Liu, Zhang, Muennighoff, Lian, and Nie]{xiao2024c}
Shitao Xiao, Zheng Liu, Peitian Zhang, Niklas Muennighoff, Defu Lian, and Jian-Yun Nie.
\newblock C-pack: Packed resources for general chinese embeddings.
\newblock In \emph{Proceedings of the 47th international ACM SIGIR conference on research and development in information retrieval}, pages 641--649, 2024{\natexlab{b}}.

\bibitem[Chen et~al.(2018)Chen, Chen, Liu, Yang, Lu, and Tang]{chen2018bq}
Jing Chen, Qingcai Chen, Xin Liu, Haijun Yang, Daohe Lu, and Buzhou Tang.
\newblock The bq corpus: A large-scale domain-specific chinese corpus for sentence semantic equivalence identification.
\newblock In \emph{Proceedings of the 2018 conference on empirical methods in natural language processing}, pages 4946--4951, 2018.

\bibitem[Liu et~al.(2018)Liu, Chen, Deng, Zeng, Chen, Li, and Tang]{liu2018lcqmc}
Xin Liu, Qingcai Chen, Chong Deng, Huajun Zeng, Jing Chen, Dongfang Li, and Buzhou Tang.
\newblock Lcqmc: A large-scale chinese question matching corpus.
\newblock In \emph{Proceedings of the 27th international conference on computational linguistics}, pages 1952--1962, 2018.

\bibitem[Yang et~al.(2019)Yang, Zhang, Tar, and Baldridge]{yang2019paws}
Yinfei Yang, Yuan Zhang, Chris Tar, and Jason Baldridge.
\newblock Paws-x: A cross-lingual adversarial dataset for paraphrase identification.
\newblock \emph{arXiv preprint arXiv:1908.11828}, 2019.

\bibitem[Cer et~al.(2017)Cer, Diab, Agirre, Lopez-Gazpio, and Specia]{cer2017semeval}
Daniel Cer, Mona Diab, Eneko Agirre, Inigo Lopez-Gazpio, and Lucia Specia.
\newblock Semeval-2017 task 1: Semantic textual similarity-multilingual and cross-lingual focused evaluation.
\newblock \emph{arXiv preprint arXiv:1708.00055}, 2017.

\end{thebibliography}

\newpage
\appendix
\section{Data Collection}

\begin{table*}[h!]
\centering
\caption{Statistics of the datasets used in our experiments. The collection includes 8 IR tasks and 7 STS tasks, covering diverse domains. Tasks marked with `-` in the \textbf{\#Train} column are evaluated in a zero-shot setting.}
\resizebox{\textwidth}{!}{%
\begin{tabular}{lcccl}
\midrule
\textbf{Name} & \textbf{Type} & \textbf{\#Train} & \textbf{\#Test} & \textbf{Description} \\
\midrule
CmedqaRetrieval \citep{qiu2022dureader_retrieval} &  Retrieval  &  99,904   &  4,000  &  Online medical consultation texts  \\
CovidRetrieval \citep{qiu2022dureader_retrieval} &  Retrieval  &  -  &  949  &  The COVID-19 news article retrieval dataset  \\
DuRetrieval \citep{qiu2022dureader_retrieval} &  Retrieval  &  83,456   &  2,000  &  A large-scale Chinese web search engine paragraph retrieval benchmark  \\
MMarcoRetrieval \citep{bonifacio2021mmarco} &  Retrieval  &  -   &  6,980  &  the multilingual version of the MS MARCO paragraph ranking dataset  \\
T2Retrieval \citep{xie2023t2ranking} &  Retrieval  &  698,752   &  22,800  &  T2Ranking: A large-scale Chinese paragraph ranking benchmark  \\
EcomRetrieval \citep{long2022multi} &  Retrieval  &  81,920   &  1,000  &  Multi-CPR: A Multi Domain Chinese Dataset for Passage Retrieval  \\
MedicalRetrieval \citep{long2022multi}  &  Retrieval  &  -   &  1,000  &  Multi-CPR: A Multi Domain Chinese Dataset for Passage Retrieval  \\
VideoRetrieval \citep{long2022multi}  &  Retrieval  &  82,560   &  1,000  &  Multi-CPR: A Multi Domain Chinese Dataset for Passage Retrieval  \\
AFQMC \citep{xu2020clue}  &  STS  &  34,334  &  3,861  &  Ant Financial Question Matching Corpus  \\
ATEC \citep{xiao2024c} &  STS  &  62,477  &  20,000  &  ATEC NLP Sentence Pair Similarity Competition  \\
BQ \citep{chen2018bq}  &  STS  &  100,000  &  10,000  &  Banking Question Semantic Similarity  \\
LCQMC \citep{liu2018lcqmc}  &  STS  &  238,766  &  12,500  &  Large-scale Chinese Question Matching Corpus  \\
PAWSX \citep{yang2019paws} &  STS   &  49,129  &  2,000 &  Translated PAWS evaluation pairs  \\
QBQTC \citep{xu2020clue} &  STS  &  180,000  &  5,000  &  QQ Browser Query Title Corpus  \\
STSB \citep{cer2017semeval}  &  STS  &  5,231  &  1,360  &  Translated STS-B into Chinese  \\
\midrule
\end{tabular}%
}
\end{table*}

\end{document}